\documentclass[runningheads]{llncs}

\usepackage{accv}

\usepackage{accvabbrv}

\usepackage{graphicx}
\usepackage{booktabs}

\usepackage[accsupp]{axessibility}  %

\usepackage[pagebackref,breaklinks,colorlinks,citecolor=accvblue]{hyperref}

\usepackage{orcidlink}
\usepackage{graphicx}
\usepackage{booktabs}
\def\method{HELIOS}
\usepackage[accsupp]{axessibility}  %

\usepackage{hyperref}
\usepackage{multirow} 
\usepackage{orcidlink}

\usepackage{overpic}

\begin{document}

\title{\method: From midnight to noon, continuous outdoor urban scene relighting} 

\titlerunning{HELIOS}

\author{
    Hala Djeghim$^
    {1,2}$ \qquad
    Nathan Piasco$^{1}$ \qquad
    Luis Roldão$^{1}$ \qquad
    Moussab Bennehar$^{1}$ \qquad 
    Dzmitry Tsishkou$^{1}$ \qquad
    Céline Loscos$^{3}$  \qquad 
  Désiré Sidibé$^{2}$ 
    \\
    }

\authorrunning{H.~Djeghim et al.}

\institute{Noah's Ark, Huawei Paris Research Center, France \and
IBISC, Universit\'e Paris-Saclay, Univ Evry, France \and   L Research, France
}

\maketitle

\begin{figure*}[h!] %
  \centering
  \includegraphics[width=1.0\textwidth]{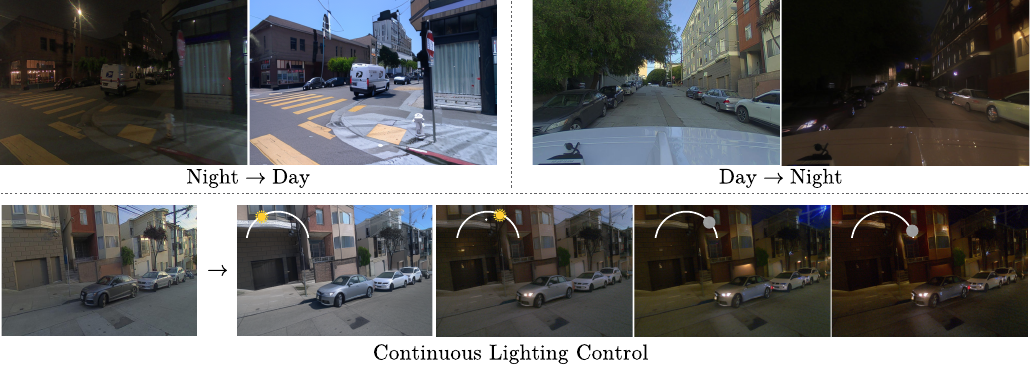} %
  \caption{\textbf{Continuous relighting control -- } From a single driving image under real-world lighting conditions, \method~produces realistic relightings for both Night-to-Day (top left), and Day-to-Night (top right) tasks while preserving the scene content. Controlled by the sun altitude, \method~enables smooth continuous relightings across the day-night cycle (bottom).}
  \label{fig:teaser}
\end{figure*}
\begin{abstract}
Modifying the illumination of driving images is a fundamental challenge, as most datasets are captured at specific times of day. Existing methods rely on synthetic data or paired multi-illumination supervision, which limits their generalization to the diverse and challenging conditions of real-world scenarios. To address this, we propose \method, a novel image relighting approach that relies on unlabeled real-world datasets without requiring any paired images for training. Our approach integrates albedo-based conditioning into a cycle-consistent diffusion pipeline to prevent identity collapse and ensure accurate domain translation. To handle low-visibility nighttime conditions, we introduce a robust albedo distillation strategy that transfers structural stability from the daytime domain. Additionally, we replace traditional text prompts with a fine-grained control mechanism based on GPS-derived solar angles, enabling smooth and continuous lighting manipulation across the day-night cycle. Through extensive evaluation and a user study, we demonstrate that \method~produces structurally consistent and realistic results in both night-to-day and day-to-night tasks, outperforming state-of-the-art methods.
\end{abstract}
    
\section{Introduction}
\label{sec:intro}

Illumination plays a crucial role in shaping the visual appearance of a scene, as it controls the perceptual interpretation of both geometry and material properties. In the context of autonomous driving, the ability to manipulate lighting while preserving structural and semantic consistency is essential for downstream tasks, such as improving perception robustness and enabling realistic scene editing for data augmentation. Indeed, most public datasets are captured at limited times of day; simulating the same scene under different lighting conditions and at various times of day is highly valuable.

Realistic relighting of driving scenes remains a highly challenging problem. It requires accurate modeling of complex light transport and surface reflectance under highly dynamic and uncontrolled outdoor conditions. Although data-driven approaches have demonstrated impressive results in controlled settings, such as indoor environments or object-centric scenarios~\cite{wang2021learningindoorinverserendering, Zhu_2023_CVPR}, their generalization to real-world driving scenes remains limited. To resolve the impossibility of capturing perfectly aligned multi-illumination datasets in the wild, many frameworks rely heavily on synthetic data. However, such synthetic priors fail to capture the full complexity of real driving environments, leading to a domain gap that significantly degrades performance.

Recent efforts have attempted to mitigate this issue through multi-stage training pipelines combining synthetic data generation and auto-labeling of real-world data~\cite{DiffusionRenderer, he2025unirelight}. Although these methods achieve good results in outdoor daytime settings, they remain fundamentally constrained by the domain gap between synthetic and real \textbf{nighttime} distributions, where sensor noise, low signal-to-noise ratios, and complex illumination effects differ significantly from synthetic data.

To address both the lack of multi-illumination supervision data and the domain gap between real and synthetic data, we propose a novel single-image relighting framework trained exclusively on real-world datasets, without requiring any paired multi-illumination data. Inspired by CycleNet~\cite{xu2023cyclenet}, our approach leverages the strong generative priors of diffusion models to perform unpaired domain translation while preserving scene content through a cycle-consistency training strategy.

We introduce an albedo estimation module and condition the relighting process on albedo maps rather than input images. Therefore, the model can easily decompose the reflectance properties of the images from its illumination. This prevents the model from falling into identity collapse and forces the model to learn illumination transfer independently from the input image. 

Furthermore, to address the ambiguity of nighttime conditions where visibility is low, we propose a robust albedo distillation strategy. This approach ensures reliable nighttime albedo prediction even when structural details are partially invisible or obscured by shadows, by transferring knowledge from a well-lit domain. 

Finally, we compute solar angles associated to each image by exploiting GPS metadata publicly available in most driving datasets. The solar angles are encoded into a continuous control signal to replace the coarse text-based prompts. Our formulation enables a fine-grained control mechanism, allowing for smooth interpolation and precise, continuous lighting manipulation.\\

\noindent In summary, our main contributions are as follows:
\begin{itemize}
    \item We propose an unpaired real-world training strategy specifically tailored for driving images relighting,
    \item We introduce albedo-based conditioning in CycleNet to prevent identity collapse and force the model to decouple reflectance from lighting during the translation process,
    \item We propose an albedo distillation strategy to solve nighttime failure cases,
     \item We replace ambiguous, discrete text prompts with a fine-grained control mechanism based on GPS-derived solar angles, enabling smooth and precise interpolation across the day-night cycle.
\end{itemize}

\section{Related work}

\subsection{Relighting via Inverse Rendering} Relighting involves modifying the illumination of an image or scene while maintaining structural consistency. Traditional techniques rely on inverse rendering to decompose a scene into its intrinsic properties (G-Buffer), including geometry, albedo, and global illumination parameters~\cite{CG}. While recent advances in rendering, such as NeRF~\cite{2020NeRF} and 3D Gaussian Splatting~\cite{Kerbl20233DGS}, have enabled high-quality relightable assets, these methods often require per-scene optimization and are typically restricted to controlled domains such as objects~\cite{gao2024relightable3dgaussiansrealistic, srinivasan2020nervneuralreflectancevisibility, Liang_2024_CVPR}, human portraits~\cite{Kanamori_2018, zheng2025groomlighthybridinverserendering}, or indoor environments~\cite{wang2021learningindoorinverserendering, Zhu_2023_CVPR}.

In the context of large-scale outdoor driving scenarios, inverse rendering remains a highly ill-posed problem. Existing solutions for urban environments~\cite{wang2023fegr, lin2025urbanirlargescaleurbanscene} often rely on manually designed priors and are effective only in high-quality daytime captures. The complexity of real-world driving datasets, characterized by transient lighting and non-uniform material properties, makes it non-trivial to disentangle illumination from reflectance. Recent data-driven approaches~\cite{Kocsis_2024_CVPR, Zeng_2024} attempt to bridge this gap by training on synthetic data or augmenting with real datasets~\cite{DiffusionRenderer}. However, these frameworks often fail to generalize to challenging nighttime images where there is limited visibility.

\subsection{Learning-based Relighting} An emerging alternative focuses on learning the relighting mapping directly via generative models~\cite{zeng2024dilightnet, NEURIPS2024_ff737391, Xing2024luminet, he2025unirelight}. These methods leverage the priors of diffusion models to synthesize relit images without the need for an intermediate G-buffer. While these models show impressive generalization on standard datasets, they struggle with the complexities of driving environments due to the absence of multi-illumination ground truth.
Notably, frameworks such as UniRelight~\cite{he2025unirelight} demonstrate high-fidelity results in daytime settings, but their performance in nighttime conditions remains undocumented. They rely on pseudo-labels generated by DiffusionRenderer~\cite{DiffusionRenderer}, which fail to produce accurate albedo maps in nighttime images.

In contrast to these methods, our approach utilizes a cycle-consistent training and a robust albedo distillation strategy, enabling direct learning of relighting from real-world driving datasets that remain structurally accurate even in nighttime conditions.

\subsection{Unpaired Image-to-Image Translation} Unpaired Image-to-Image (I2I) translation aims to learn a mapping between two distinct domains without pixel-aligned supervision. CycleGAN~\cite{CycleGAN2017} established the cycle-consistency constraint to preserve structural content across domain shifts. While effective, GAN-based methods often suffer from limited diversity and instability in complex scenes. Recent works have integrated cycle-consistency into the Latent Diffusion pipeline; notably, CycleNet~\cite{xu2023cyclenet} leverages the generative power of Stable Diffusion to achieve high-fidelity translation. 

However, CycleNet~\cite{xu2023cyclenet} is limited by a trade-off between structural consistency and domain translation, which often leads to identity collapse in complex tasks such as driving images relighting.
Our work builds upon the cycle-consistent diffusion training strategy but addresses its limitations by replacing input image conditioning with a robust albedo prior and introducing a continuous relighting control.

\begin{figure*}[t]
    \centering
    \begin{overpic}[width=1\textwidth]{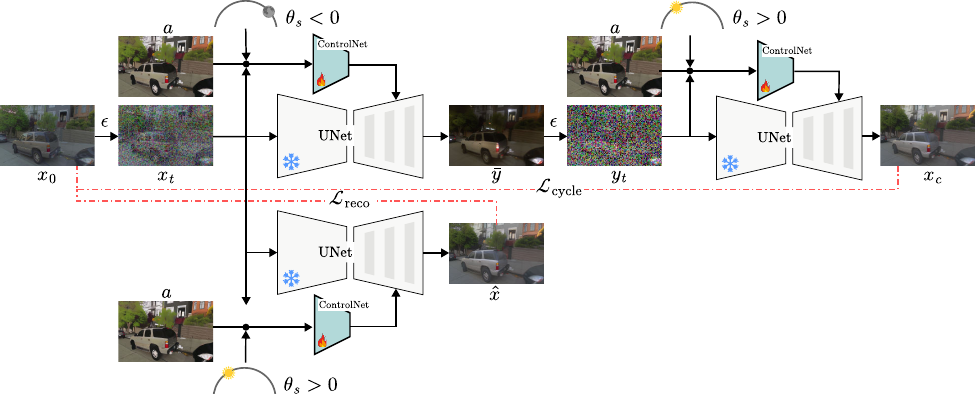}
    \end{overpic}
    \caption{\textbf{Overview of the HELIOS Architecture.} Our framework performs unpaired relighting conditioned on the albedo $a$ and solar altitude $\theta_s$. Starting from a daytime source sample $x_0$, the model first optimizes a reconstruction objective $\mathcal{L}_{\text{reco}}$ conditioned on the elevation $\theta_s > 0$. Domain translation is achieved by generating a relit nighttime representation $\bar{y}$ conditioned on $\theta_s < 0$. To ensure structural integrity, $\bar{y}$ is mapped back to the source domain as $x_c$ via the cycle-consistency loss $\mathcal{L}_{cycle}$.}
    \label{fig:method-architecture}
    \hfill
\end{figure*}

\section{Background}

\subsection{CycleNet}
\label{background_cyclenet}
Unpaired I2I translation aims to learn a mapping between domains $\mathcal{X}$ and $\mathcal{Y}$ without any aligned paired samples. CycleNet~\cite{xu2023cyclenet} adapts the cycle-consistency principle~\cite{CycleGAN2017} to a diffusion framework by enforcing that an image $x \in \mathcal{X}$, when translated to domain $\mathcal{Y}$, subsequently translated back, remains reconstructible to the original source.

This relies on the assumption that the structural condition $c_{\text{img}}$ remains invariant, while the target domain is controlled by the conditioning signal $c_{\text{text}}$. To implement this, they use a ControlNet~\cite{zhang2023adding} architecture. The network $F_\theta$ is trained to predict the clean sample $x_0$ at a specific timestep $t$ for a given noisy latent $x_t$.
They first introduce a \textbf{reconstruction loss} that ensures the model can recover the original input when conditioned on its own domain label: 

\begin{equation}
\mathcal{L}_{\text{x}\to\text{x} } =  \mathbb{E}_{x, \epsilon \sim \mathcal{N}(0, I), t} \left[ \left\| x_0 - F_\theta(x_t, c_{x}, x_0) \right\|_2^2 \right],
\label{eq:noise-pred}
\end{equation}

They define a \textbf{cycle consistency loss} that ensures that translation from both domains can reconstruct the original image $x_0$. Let $\bar{y} = F_\theta(x_t, c_y, x_0)$ be an intermediate translation where $y_t = \bar{y} + \epsilon_y$ is its noised version. The cycle loss is then defined as: 

\begin{equation}
\mathcal{L}_{\text{x}\to\text{y}\to\text{x}} = \mathbb{E}_{x, \epsilon \sim \mathcal{N}(0, I), t} \left[ \left\| x_0 - F_\theta(y_t, c_{x}, x_0) \right\|_2^2 \right],
\label{eq:noise-pred}
\end{equation}

Finally, an \textbf{invariance loss} is used to ensure the target domain remains stable under repeated transfers:

\begin{equation}
\mathcal{L}_{\text{x}\to\text{y}\to\text{y}} =\mathbb{E}_{z, \epsilon \sim \mathcal{N}(0, I), t} \left[ \left\| F_\theta(x_t, c_{y}, x_0)  - F_\theta(x_t, c_{y}, \bar{y}) \right\|_2^2 \right].
\label{eq:inv_loss}
\end{equation}

\section{Method}

Our goal is to perform unpaired image-to-image translation to relight driving scenes from a single image. To achieve this, our method is trained exclusively on unpaired real-world datasets, without any synthetic supervision or paired captures.
 
We first introduce an albedo regularization strategy in Sec.~\ref{subsec:albedo}, replacing image conditioning with an invariant albedo prior to prevent identity collapse. In Sec.~\ref{subsec:continuous} we present a continuous relighting control.
Finally, in Sec.~\ref{sec:albedo-distillation}, we propose a novel distillation strategy to train a domain-invariant albedo estimator. This ensures reliable structural preservation and consistent predictions even in low-visibility nighttime conditions.

\subsection{HELIOS}

We formulate our relighting task as an unpaired translation problem using the cycle-consistency training defined in Sec.~\ref{background_cyclenet}. 

We aim to learn a function $F_\theta$ capable of mapping an image $z_0$ from a source distribution to a target distribution. We define $\mathcal{X}$ and $\mathcal{Y}$ as the daytime and nighttime driving domains, respectively. 

The text conditioning $c_{\text{text}} \in \{c_x, c_y\}$ represents the target domain ( `day' or `night'), while the structural conditioning $c_{\text{img}} \in \{x_0, y_0\}$ provides the spatial context.

\begin{figure*}[tb] 
    \centering
    \def\fgsize{0.18}
    \scriptsize
    \setlength{\tabcolsep}{0.0035\linewidth}
    \renewcommand{\arraystretch}{1.0}
    \begin{tabular}{c @{\hspace{7pt}} ccccc}%
         Input  & Epoch 1 &  Epoch 2 & Epoch 3 & Epoch 4 \\ 
        \includegraphics[clip=false, trim={0 0 0 0},width=\fgsize\columnwidth]{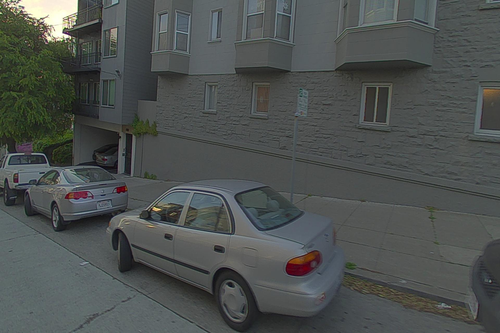} & 
        \includegraphics[clip=false, trim={0 0 0 0},,width=\fgsize\columnwidth]{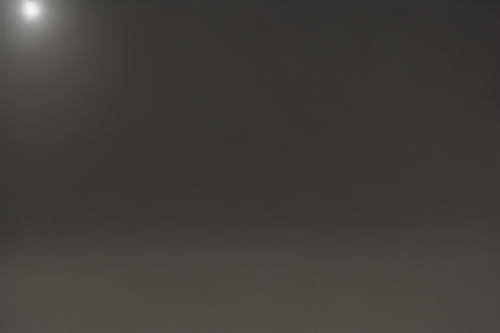} & 
        \includegraphics[clip=false, trim={0 0 0 0},,width=\fgsize\columnwidth]{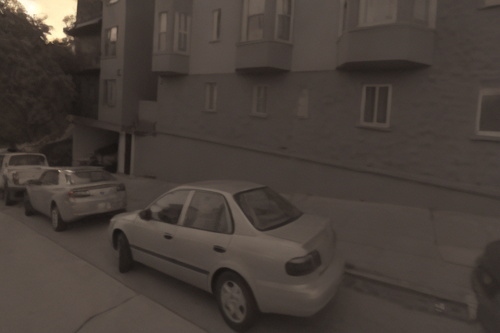} & 
        \includegraphics[clip=false, trim={0 0 0 0},,width=\fgsize\columnwidth]{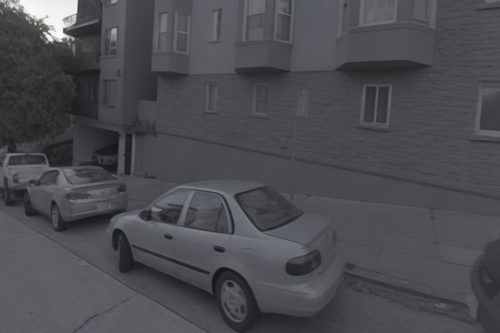} & 
        \includegraphics[clip=false, trim={0 0 0 0},,width=\fgsize\columnwidth]{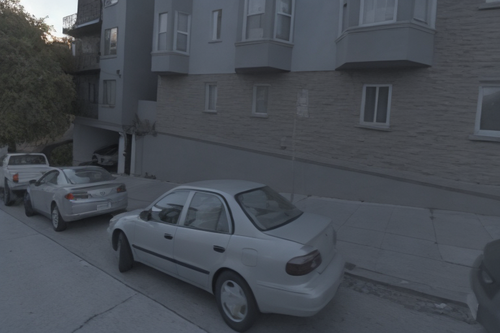} &            
    \end{tabular}
    \caption{  \textbf{CycleNet~\cite{xu2023cyclenet} failure case for Day-to-Night relighting.} We show the input, and predictions at different epochs. CycleNet fails to ensure a proper trade-off between translation and structural consistency. At the early stage, the model attempts translation to the night domain but suffers from an identity collapse, where it copies the input image without any meaningful nighttime lighting effects.}

    \label{fig:failure-cyclenet}

\end{figure*}

A fundamental limitation of cycle-consistency, as noted in ACLGAN~\cite{Zhao2020ACLGAN} and CycleNet~\cite{xu2023cyclenet} (in \textit{Sec 5.4}), 
is the need for the translated image $\bar{y}$ to retain sufficient information to allow for a perfect reconstruction of the source $x_0$.
When the model is conditioned on the input image ($c_{\text{img}} = x_0$), this constraint becomes overly restrictive. It prevents a proper trade-off between structural consistency and domain translation. 
In the complex context of driving scenes this constraint often prevents the model from performing a significant lighting shift while maintaining the scene content. 
We demonstrate this failure case in Fig.~\ref{fig:failure-cyclenet}. In early training phases, the model attempts translation but fails to preserve the underlying structure. As training converges, the model eventually collapses toward an identity mapping, where $ F_\theta(x_t, c_y, x_0) \approx x_0$. This satisfies the reconstruction objective but fails to achieve any meaningful relighting.

\subsection{Light invariant prior as conditioning}
\label{subsec:albedo}

To address the constraints in standard cycle-consistent frameworks, we introduce an Albedo Estimator $\mathcal{A}$ that extracts the intrinsic reflectance $a = \mathcal{A}(x_0)$ from the source image. The albedo represents the lighting-invariant component of the image, which captures the base color without any lighting effects. 

The extracted albedo is then used as a conditioning signal to replace the input image $c_{img}$. 
This prevents the model from relying on the conditioning image to satisfy the cycle-consistency objective. This is possible because the albedo $a$ remains constant across domain translations of the same image, such that the reflectance of the source $x_0$ is identical to its translated counterpart $\bar{y}$.
By leveraging this invariant prior, the optimization problem is simplified; the model no longer requires an explicit invariance loss to stabilize the target domain. Instead, the network learns to map the illumination dictated by the semantic conditioning $c_{text}$ onto the structural conditioning provided by $a$. 

The final training objectives are redefined as:

\begin{equation}
\begin{split}
    \mathcal{L}_{\text{reco}} = \mathbb{E}_{x, \epsilon \sim \mathcal{N}(0, I), t} \left[ \left\| x_0 - F_\theta(x_t, c_{x}, a) \right\|_2^2 \right] \\
\end{split}
\label{eq:reco}
\end{equation}

\begin{equation}
\mathcal{L}_{\text{cycle}} = \mathbb{E}_{x, \epsilon \sim \mathcal{N}(0, I), t} \left[ \left\| x_0 - F_\theta(y_t, c_{x}, a) \right\|_2^2 \right].
\label{eq:cycle}
\end{equation}

\subsection{Continuous relighting}
\label{subsec:continuous}
To resolve the ambiguity of text prompts that fail to capture subtle and gradual transitions of outdoor lighting, we replace textual prompts with a continuous conditioning mechanism based on the solar altitude angle $\theta_s$. For any given sample, $\theta_s$ represents the elevation of the sun relative to the observer's local horizon:

\begin{equation} \theta_s = \arcsin(\sin(\phi) \sin (\delta) + \cos (\phi) \cos (\delta) \cos (h)), \label{eq:solar_alt_final} \end{equation}

\noindent where $\phi$ is the geographic latitude, $\delta$ is the solar declination, and $h$ is the local hour angle. These parameters are extracted from GPS metadata for each training sample.

Under this convention, $\theta_s > 0^\circ$ represents daytime conditions (sun above the horizon), while $\theta_s < 0^\circ$ corresponds to nighttime states. The scalar $\theta_s$ is mapped into a high-dimensional feature space via positional encoding and injected into the cross-attention layers of the UNet, replacing the text prompt. This continuous formulation enables the model to learn smooth illumination states, allowing for fine-grained control over the global lighting and the synthesis of realistic transitions along the sun's trajectory.\\

\section{Robust Night-to-Day Translation via  Albedo Distillation}
\label{sec:albedo-distillation}
\begin{figure*}[tb] 
\centering
\scriptsize
\setlength{\tabcolsep}{0.005\linewidth} %
\renewcommand{\arraystretch}{1.0}

\begin{tabular}{ccc@{\hspace{10pt}}ccc}
    & Input & Albedo & & Input & Albedo \\ 

    \rotatebox{90}{\hspace{15pt}Day} &
    \includegraphics[width=0.22\columnwidth]{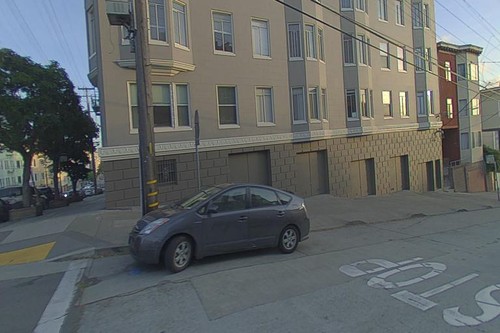} & 
    \includegraphics[width=0.22\columnwidth]{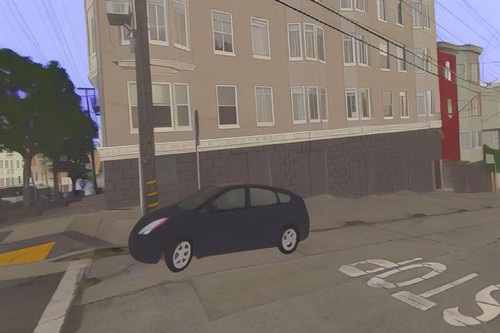} &
    
    \rotatebox{90}{\hspace{12pt}Night} &
    \includegraphics[width=0.22\columnwidth]{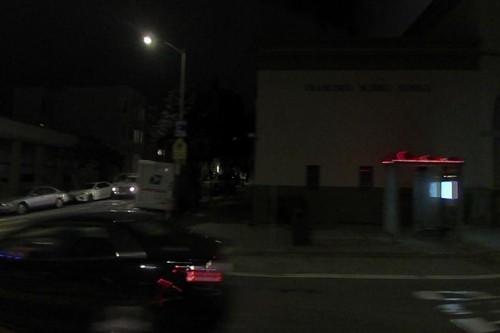} & 
    \includegraphics[width=0.22\columnwidth]{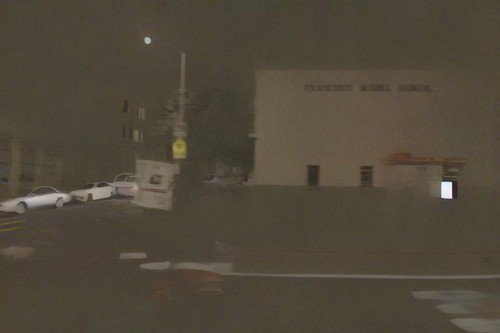} \\
\end{tabular}

\caption{\textbf{Failure cases of DiffusionRenderer~\cite{DiffusionRenderer} in nighttime illumination.} While the baseline provides accurate albedo estimates for daytime images (left), it suffers from significant structural loss and color drifting under nocturnal conditions (right).}
\label{fig:albedos_comparison}
\end{figure*}

While we can use any state-of-the-art intrinsic decomposition method as the albedo estimator $\mathcal{A}$ introduced in Sec~\ref{subsec:albedo}, existing approaches such as DiffusionRenderer~\cite{DiffusionRenderer} often fail under nighttime driving conditions.
Specifically, in areas obscured by deep shadows, these estimators struggle to maintain consistency, resulting in incomplete albedo maps, as shown in Fig.~\ref{fig:albedos_comparison}.
These degraded reflectance maps lead to structural failures during translation, where the model fails to recover geometry obscured in the nighttime input. We address this by proposing a synthetic distillation strategy to train a domain-invariant albedo estimator $\mathcal{A}$ that maintains structural consistency across all illumination states.

\subsection{Cross-Domain Albedo Distillation}
\label{subsec:robust-albedo}

Based on the observation that intrinsic decomposition is significantly more stable for daytime images $\mathcal{X}$ compared to nighttime images $\mathcal{Y}$, we introduce a multi-stage process to distill daytime knowledge into a robust nighttime and daytime albedo estimator.

\paragraph{Synthetic Illumination Sampling.} We use an initial Day-to-Night (D2N) \method~model to generate a multi-illumination dataset. 
For each daytime image $x_0$, we produce a sequence of relit variants $\{\bar{y}_i\}_{i=1}^3$ and $\{\bar{x}_i\}_{i=1}^2$ (three nighttime and two daytime). Since the scene geometry remains static across these variants, the daytime computed albedo $a$ serves as a pseudo-ground-truth for all generated illumination.
\paragraph{Robust Albedo Estimator.} We fine-tune an image-conditioned Stable Diffusion model to map any image $x \in \{\mathcal{X} \cup \bar{\mathcal{Y}}\}$ to its invariant albedo $a$. The estimator $\mathcal{A}$ is optimized via a latent-space denoising objective:

\begin{equation}
\mathcal{L}_{\text{albedo}} = \mathbb{E}_{a, \epsilon_a \sim \mathcal{N}(0, I), t} \left[ \left\| a_0 - \mathcal{A}(a_t, x) \right\|_2^2 \right],
\label{eq:albedo}
\end{equation}

\noindent where $a_t$ is the noisy latent of the albedo map. By training on both real daytime and generated nighttime images, $\mathcal{A}$ learns to disentangle illumination from the base color, providing a robust structural prior that is domain-invariant to both daytime and nighttime conditions.

\subsection{Day-to-Night and Night-to-Day training}
The frozen robust albedo estimator is then used to re-process our entire driving dataset to generate a consistent albedo set. These maps provide a stable, light-invariant structural condition for our final unified relighting framework. An overview of our training pipeline can be found in Fig. ~\ref{fig:method-architecture}.

The final \method~model is trained using the unified objective for both Day-to-Night and Night-to-Day tasks:

\begin{equation}
\mathcal{L} = \mathcal{L}_{\text{reco}} + \mathcal{L}_{\text{cycle}},
\end{equation} 

\noindent where the albedo $a$ used as condition in Eq.~\ref{eq:reco} and ~\ref{eq:cycle} is now estimated with our robust albedo estimator introduced in Sec.~\ref{subsec:robust-albedo}. By conditioning on these distilled reflectance priors, the model effectively bypasses the ambiguities of nighttime scenes, allowing for high-fidelity Night-to-Day and Day-to-Night translations that preserve scene geometry while accurately synthesizing global illumination.

\section{Experiments}

\subsection{Experiments details}
\paragraph{Implementation details.}
We first perform a domain-specific fine-tuning on a pre-trained Stable Diffusion V2.1~\cite{Rombach_2022_CVPR} U-Net using driving images. This stage ensures that the model learns the distinct lighting distributions of both daytime and nighttime environments. Rather than using standard text-to-image supervision, we replace text-embedding conditioning with our continuous solar altitude angle $\theta_s$. We fine-tune the model with a batch size of 4 and for 100k steps, using a fixed learning rate of $1\times10^{-5}$ and the Adam optimizer. This initial training phase required approximately 24 hours.
We then train ControlNet~\cite{zhang2023adding}, initializing its weights from the fine-tuned U-Net. We keep the U-Net and VAE frozen. The model is trained with a batch size of 1 and for 300k steps using a fixed learning rate of $1\times10^{-5}$, requiring approximately four days of training. All experiments were conducted on a GPU equivalent to the NVIDIA RTX 4090. Images were resized, preserving the original aspect ratio with a minimum height or width of 512.

\paragraph{Datasets.}
We train \method~on a subset of three public driving datasets: nuScenes~\cite{nuscenes}, Waymo~\cite{Waymo} and Pandaset~\cite{pandaset}. The datasets capture a wide range of temporal conditions, providing diverse solar altitude angles $\theta_s$ (we refer to the supplementary materials for data distribution). We use a balanced distribution of daytime and nighttime sequences, for a total of 41k images. To train the albedo estimator, we use a separate subset of 6798 original images. For each of these images, we generate six different illumination versions, resulting in a multi-illumination synthetic dataset of approximately 40k images.

\paragraph{Baselines.}
We compare our method against text-guided editing models InstructPix2Pix~\cite{brooks2022instructpix2pix} and Qwen-Image~\cite{wu2025qwenimagetechnicalreport}. We also evaluate against DiffusionRenderer~\cite{DiffusionRenderer}, a relighting framework trained on paired synthetic and augmented real-world data controlled via environment maps, and CycleNet~\cite{xu2023cyclenet}, a cycle-consistent diffusion framework trained on unpaired data.
\paragraph{Metrics.}
To evaluate structural preservation, we compute DINO-struct ($\times 100$)~\cite{Tumanyan_2022_CVPR} by measuring feature-space distances between the source and relighted images, and report mIoU between segmentation maps of source and relit images as a complementary measure of structural consistency. Image quality and distribution alignment with the target domain are evaluated with the FID score~\cite{gans-fid}. For semantic accuracy, we calculate the CLIP-Score~\cite{clip} between generated outputs and text prompts corresponding to the target lighting condition. We additionally report downstream object detection performance as an image enhancement application, with results detailed in the applications section. Finally, we conduct a human perceptual study to quantify realism and preference. We report the Win rate, defined as the percentage of trials where a method was preferred by participants, and the Fail rate, representing the frequency at which outputs were rejected due to insufficient quality.

\subsection{Qualitative results}
\begin{figure*}[tb] 
\centering
\def\fgsize{0.16}
\scriptsize
\setlength{\tabcolsep}{0.0015\linewidth}
\renewcommand{\arraystretch}{1.0}
\begin{tabular}{ccccccc}%
          Input & CycleNet~\cite{xu2023cyclenet} &IP2P~\cite{brooks2022instructpix2pix} & Qwen-Img.~\cite{wu2025qwenimagetechnicalreport} & DiffRendr.~\cite{DiffusionRenderer} &\method~(ours)  \\ 

    \includegraphics[clip=false, trim={0 0 0 0},width=\fgsize\columnwidth]{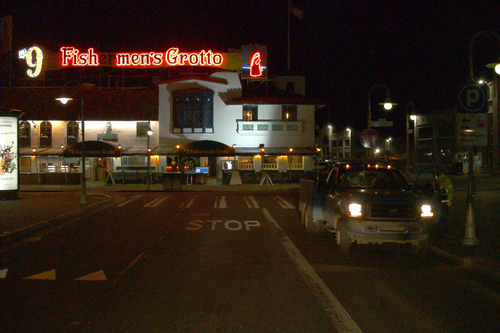} & 
    \includegraphics[clip=false, trim={0 0 0 0},width=\fgsize\columnwidth]{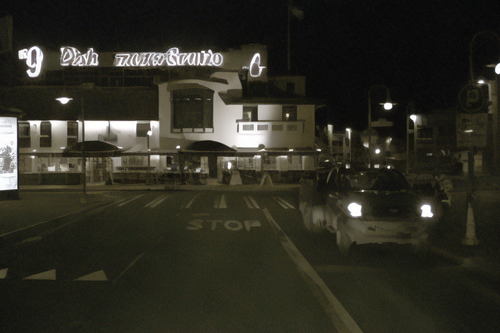} & 
    \includegraphics[clip=false, trim={0 0 0 0},,width=\fgsize\columnwidth]{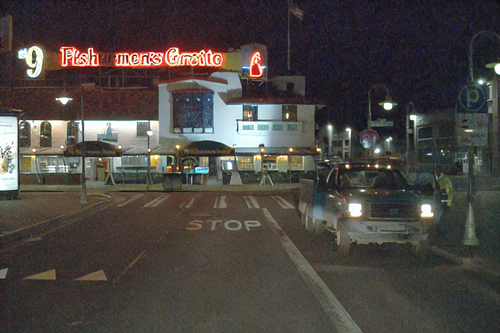} & 
    \includegraphics[clip=false, trim={0 0 0 0},,width=\fgsize\columnwidth]{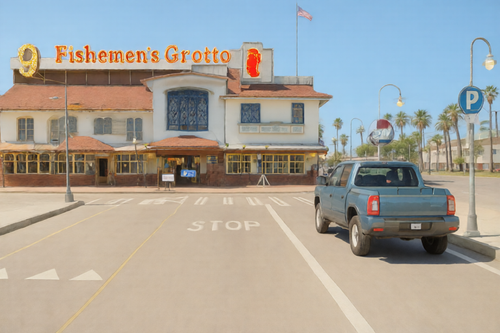} & 
     \includegraphics[clip=false, trim={0 0 0 0},,width=\fgsize\columnwidth]{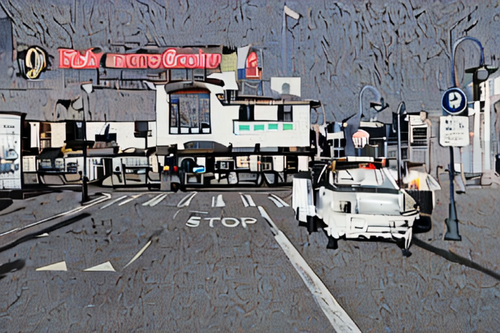} & 
    \includegraphics[clip=false, trim={0 0 0 0},,width=\fgsize\columnwidth]{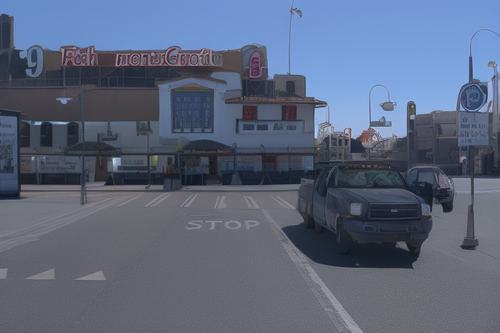} & \\
    
    \includegraphics[clip=false, trim={0 0 0 0},width=\fgsize\columnwidth]{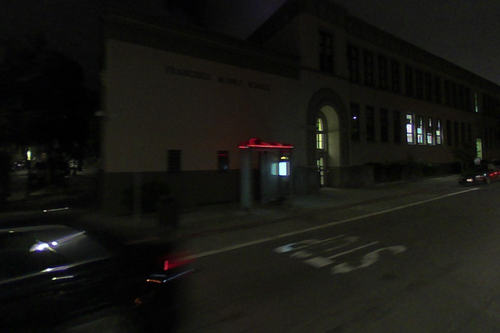} & 
    \includegraphics[clip=false, trim={0 0 0 0},width=\fgsize\columnwidth]{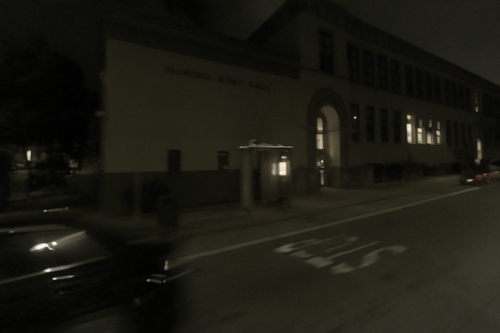} & 
    \includegraphics[clip=false, trim={0 0 0 0},,width=\fgsize\columnwidth]{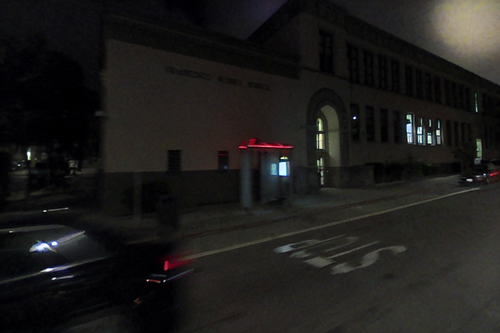} & 
    \includegraphics[clip=false, trim={0 0 0 0},,width=\fgsize\columnwidth]{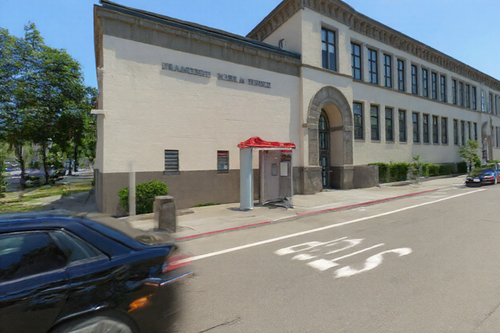} & 
     \includegraphics[clip=false, trim={0 0 0 0},,width=\fgsize\columnwidth]{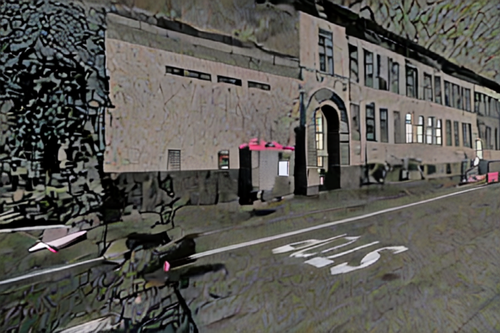} & 
    \includegraphics[clip=false, trim={0 0 0 0},,width=\fgsize\columnwidth]{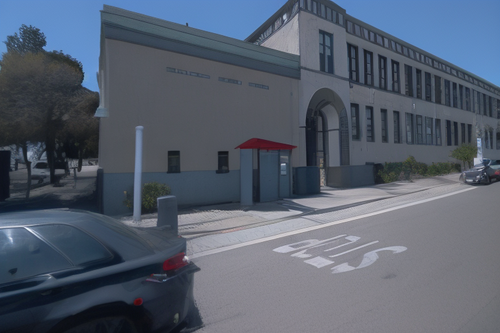} & \\
    
    \includegraphics[clip=false, trim={0 0 0 0},width=\fgsize\columnwidth]{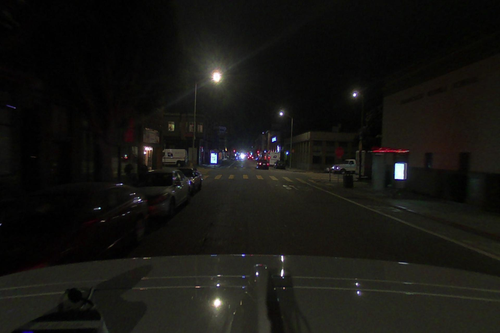} & 
    \includegraphics[clip=false, trim={0 0 0 0},width=\fgsize\columnwidth]{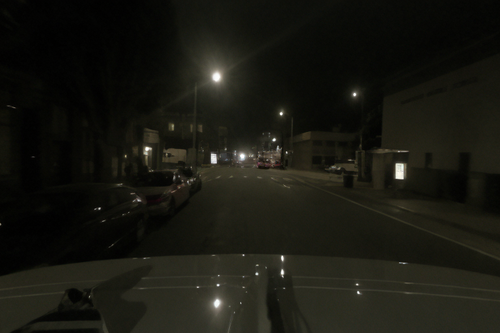} & 
    \includegraphics[clip=false, trim={0 0 0 0},,width=\fgsize\columnwidth]{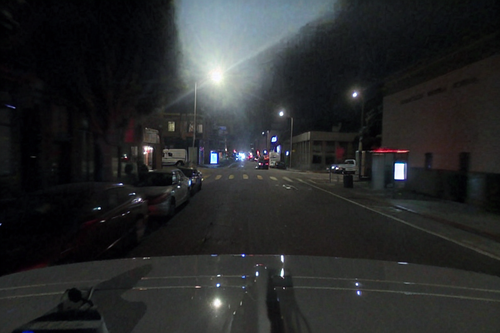} & 
    \includegraphics[clip=false, trim={0 0 0 0},,width=\fgsize\columnwidth]{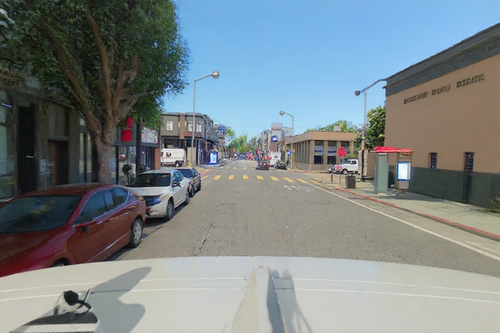} & 
     \includegraphics[clip=false, trim={0 0 0 0},,width=\fgsize\columnwidth]{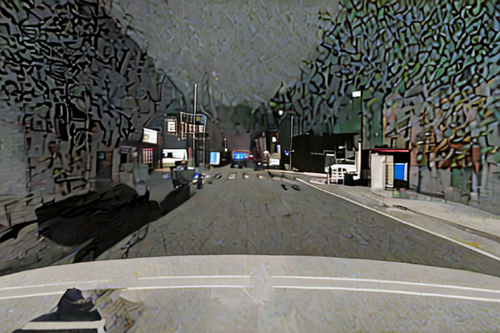} & 
    \includegraphics[clip=false, trim={0 0 0 0},,width=\fgsize\columnwidth]{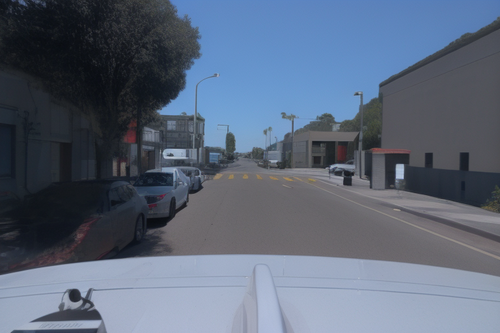} & \\
   
\end{tabular}
    \caption{\textbf{Qualitative comparison for Night-to-Day relighting.} \method~demonstrates superior performance in generating daytime illuminations, effectively recovering structural details from nighttime inputs. While Qwen-Image achieves realistic relightings as it was trained on more data with a bigger model size, it still fails to respect structure and hallucinates added objects.}
\label{fig:N2D}

\end{figure*}

\begin{figure*}[tb] 
    \centering
    \def\fgsize{0.16}
    \scriptsize
    \setlength{\tabcolsep}{0.0015\linewidth}
    \renewcommand{\arraystretch}{1.0}
    \begin{tabular}{ccccccc}%
          Input & CycleNet~\cite{xu2023cyclenet} &IP2P~\cite{brooks2022instructpix2pix} & Qwen-Img~\cite{wu2025qwenimagetechnicalreport} & DiffRendr~\cite{DiffusionRenderer} &\method~(ours)  \\ 
    
        \includegraphics[clip=false, trim={0 0 0 0},width=\fgsize\columnwidth]{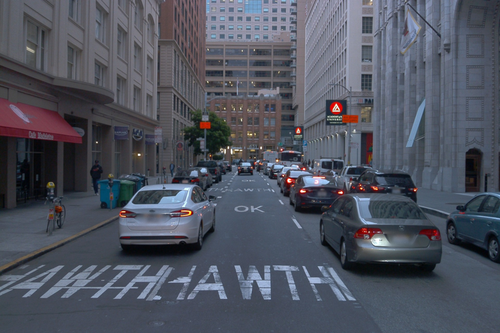} & 
        \includegraphics[clip=false, trim={0 0 0 0},width=\fgsize\columnwidth]{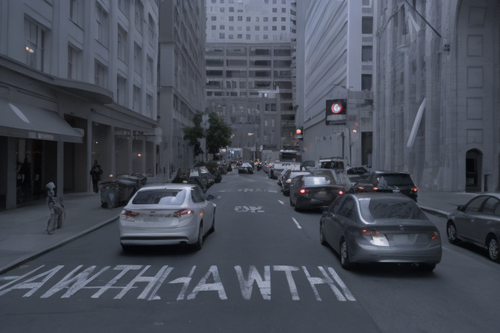} & 
        \includegraphics[clip=false, trim={0 0 0 0},,width=\fgsize\columnwidth]{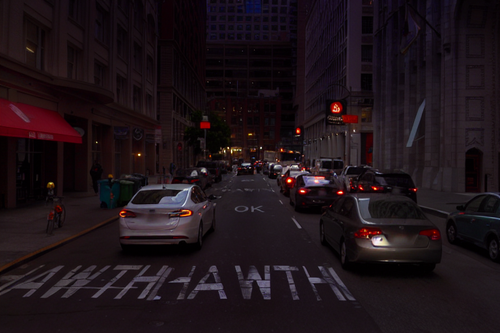} & 
        \includegraphics[clip=false, trim={0 0 0 0},,width=\fgsize\columnwidth]{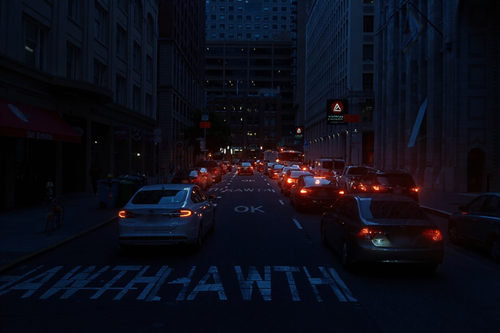} & 
         \includegraphics[clip=false, trim={0 0 0 0},,width=\fgsize\columnwidth]{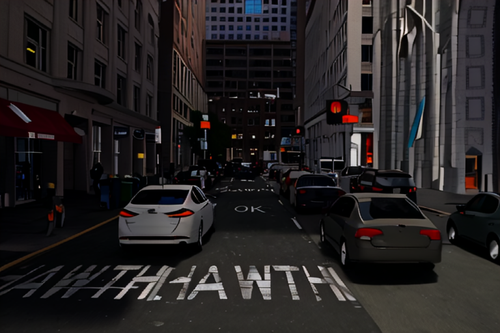} & 
        \includegraphics[clip=false, trim={0 0 0 0},,width=\fgsize\columnwidth]{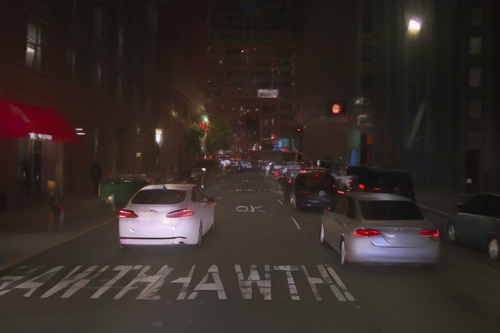} \\
        
        \includegraphics[clip=false, trim={0 0 0 0},width=\fgsize\columnwidth]{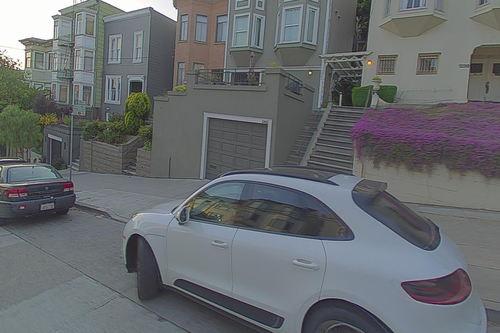} & 
        \includegraphics[clip=false, trim={0 0 0 0},width=\fgsize\columnwidth]{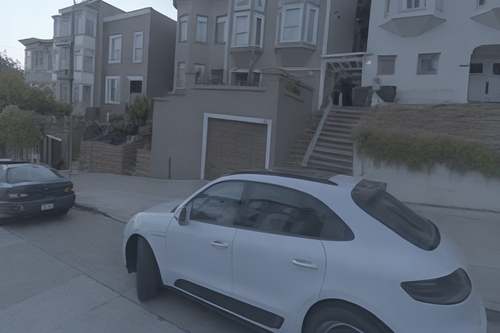} & 
        \includegraphics[clip=false, trim={0 0 0 0},,width=\fgsize\columnwidth]{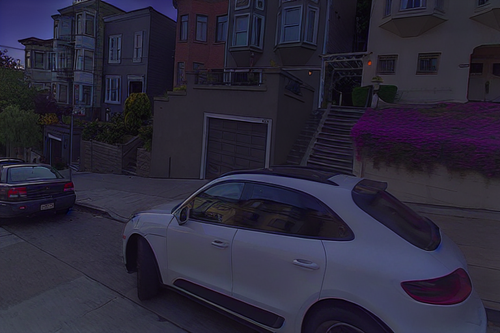} & 
        \includegraphics[clip=false, trim={0 0 0 0},,width=\fgsize\columnwidth]{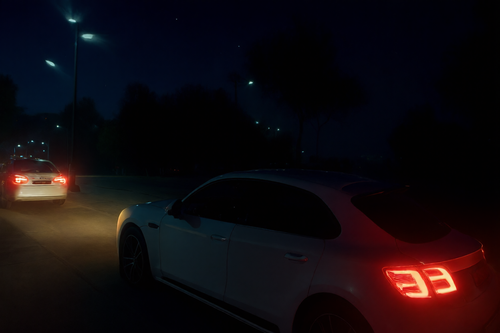} & 
         \includegraphics[clip=false, trim={0 0 0 0},,width=\fgsize\columnwidth]{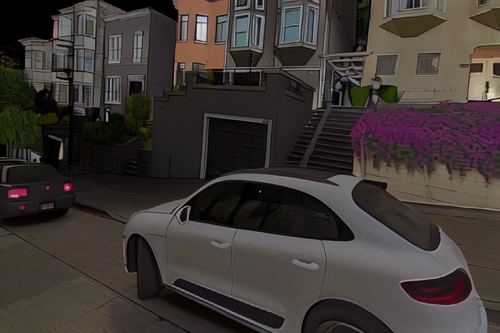} & 
        \includegraphics[clip=false, trim={0 0 0 0},,width=\fgsize\columnwidth]{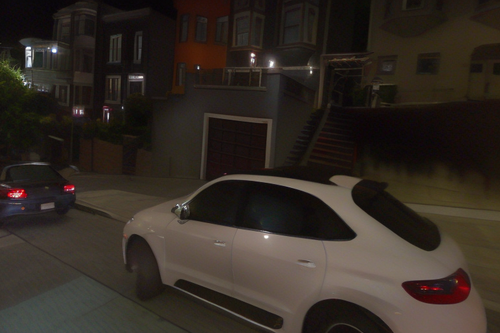} & \\
        
        \includegraphics[clip=false, trim={0 0 0 0},width=\fgsize\columnwidth]{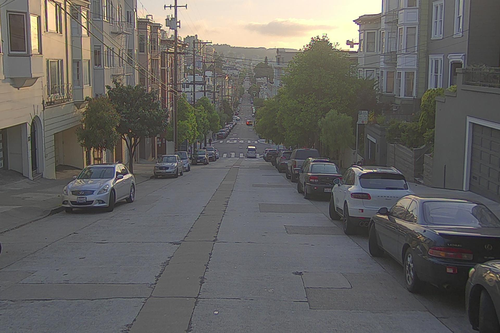} & 
        \includegraphics[clip=false, trim={0 0 0 0},width=\fgsize\columnwidth]{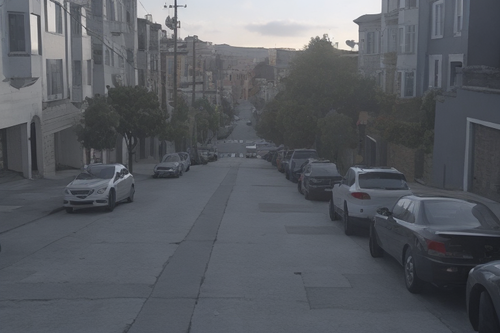} & 
        \includegraphics[clip=false, trim={0 0 0 0},,width=\fgsize\columnwidth]{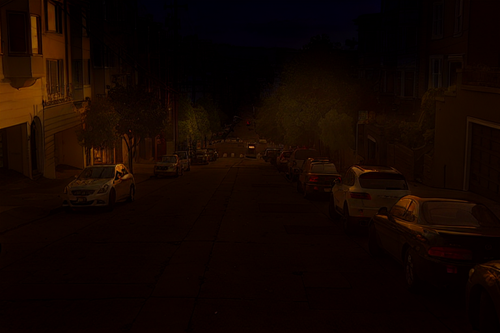} & 
        \includegraphics[clip=false, trim={0 0 0 0},,width=\fgsize\columnwidth]{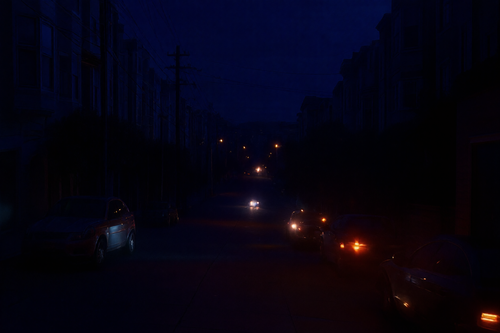} & 
         \includegraphics[clip=false, trim={0 0 0 0},,width=\fgsize\columnwidth]{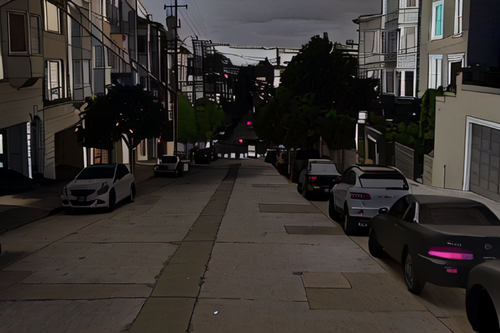} & 
        \includegraphics[clip=false, trim={0 0 0 0},,width=\fgsize\columnwidth]{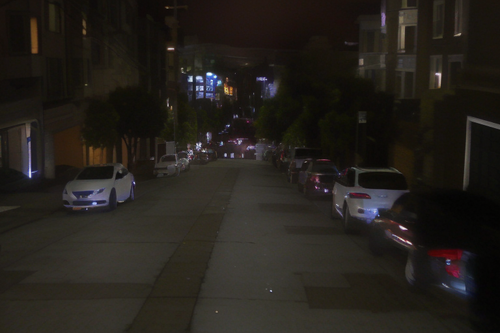} & \\
       
    \end{tabular}
    \caption{\textbf{Qualitative comparison for Day-to-Night relighting.} \method~demonstrates superior performance in generating realistic nighttime global and local illuminations while keeping the structure of the input image.}
    \label{fig:D2N}
\end{figure*}

We show in Fig.~\ref{fig:N2D} and Fig.~\ref{fig:D2N} qualitative results for Night-to-Day and Day-to-Night tasks, respectively. CycleNet consistently produces an identity image without performing domain translation. InstructPix2Pix, while struggling to achieve Night-to-Day relighting, produces global night lighting but omits local lighting effects.
DiffusionRenderer is limited by strictly keeping the albedo and just adding an environment map, which results in synthetic-looking outputs. Qwen-Image shows very impressive and realistic results for Night-to-Day; however, while it achieves good global lighting for night translation, it often loses structural consistency.
In contrast,~\method~provides consistent results across both tasks, maintaining structural integrity while successfully translating both global and local lighting effects.

\begin{figure*}[t]
    \centering
    \def\fgsize{0.155}
    \scriptsize %
    \setlength{\tabcolsep}{0.5pt} %
    \renewcommand{\arraystretch}{0.8} %
    
    \begin{tabular}{c @{\hspace{4pt}} cccccc}

        & $\theta_s = 37^\circ$ & $\theta_s = 1^\circ$ & $\theta_s = -3^\circ$ & $\theta_s = -21^\circ$ & $\theta_s = -40^\circ$ \\
        
        &
        \includegraphics[width=\fgsize\linewidth]{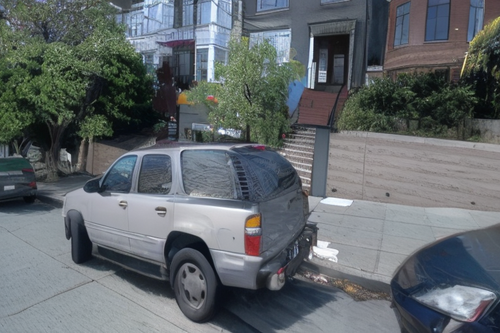} &
        \includegraphics[width=\fgsize\linewidth]{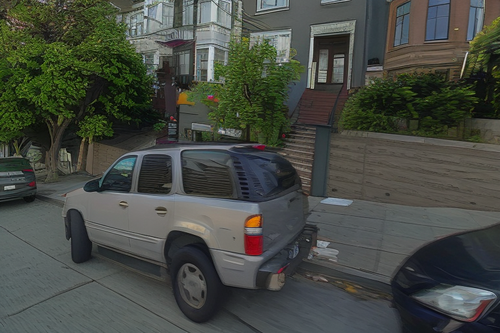} &
        \includegraphics[width=\fgsize\linewidth]{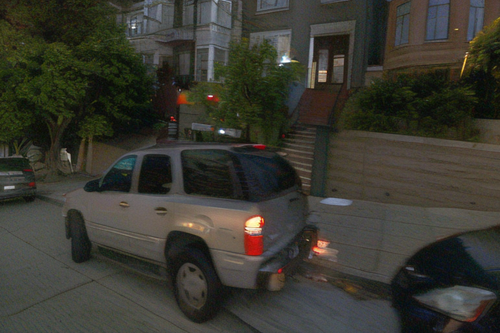} &
        \includegraphics[width=\fgsize\linewidth]{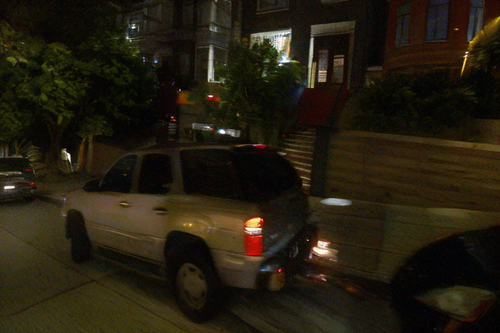} &
        \includegraphics[width=\fgsize\linewidth]{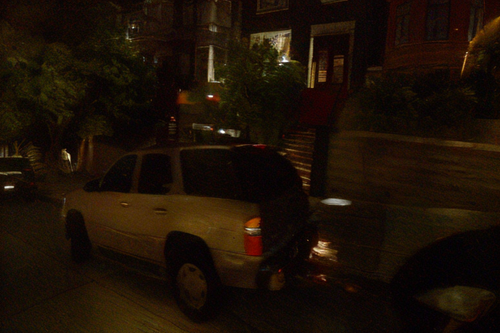} &
        \rotatebox{90}{~\method} \\

        \includegraphics[width=\fgsize\linewidth]{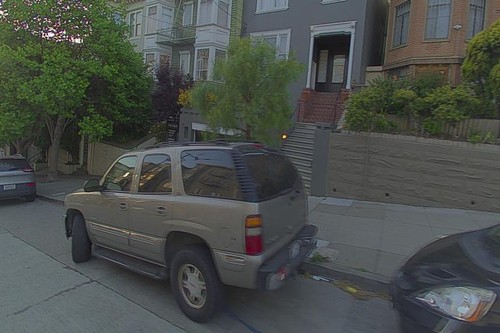} & 
        \includegraphics[width=\fgsize\linewidth]{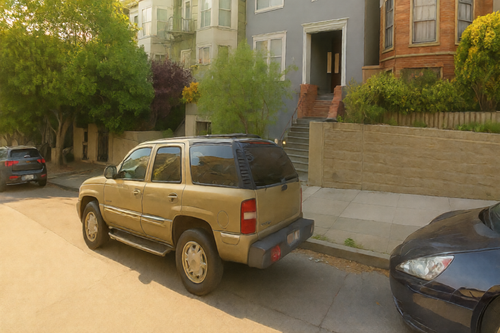} &
        \includegraphics[width=\fgsize\linewidth]{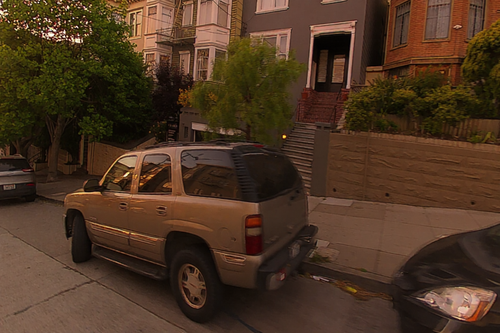} &
        \includegraphics[width=\fgsize\linewidth]{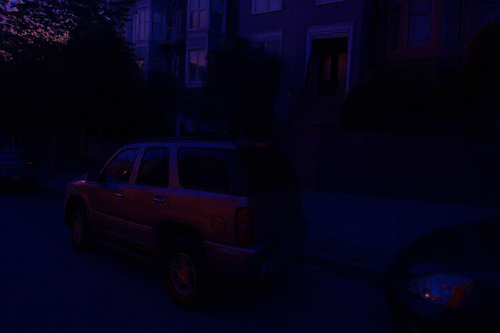} &
        \includegraphics[width=\fgsize\linewidth]{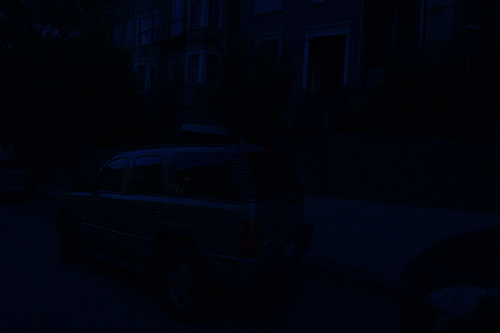} &
        \includegraphics[width=\fgsize\linewidth]{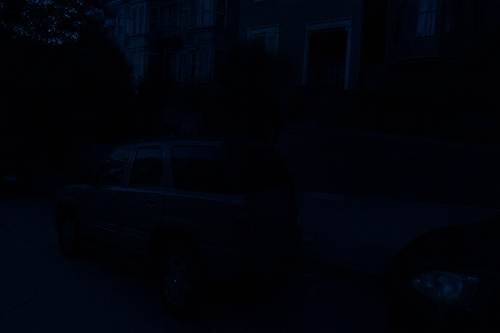} &
        \rotatebox{90}{Qwen~\cite{wu2025qwenimagetechnicalreport}} \\ 
        
        Input & Afternoon & Sunset & Twilight & Night & Deep Night \\
    \end{tabular}
    
    \vspace{-2mm} %
    \caption{\textbf{Continuous Relighting.} Our continuous solar altitude parameter ($\theta_s$) enables fine-grained, physically plausible transitions of global illumination.}
    \label{fig:continuous_transition}
    \vspace{-2mm} %
\end{figure*}

Figure ~\ref{fig:continuous_transition} compares our continuous conditioning over discrete text prompts. We refer to the supplementary material for videos that better illustrate this continuous, fine-grained control. Qwen-Image~\cite{wu2025qwenimagetechnicalreport} generates exaggerated afternoon and sunset lightings, hallucinating an intense, localized light source. In contrast, our approach generates realistic ambient illumination that preserves the natural appearance of the image. Furthermore, our continuous solar elevation parameter $\theta_s$ provides fine-grained control over low-light conditions. While discrete text models struggle to differentiate between ambiguous dark states, our method clearly isolates the distinct photometric properties of twilight, night, and deep night.

\begin{table}[htb]
\centering
\resizebox{\linewidth}{!}{%
\begin{tabular}{l cccc cccc cccc}
\cmidrule(l{1pt}r{1pt}){2-13}
& \multicolumn{4}{c}{\textbf{Night$\to$Day}} & \multicolumn{4}{c}{\textbf{Day$\to$Night}} & \multicolumn{4}{c}{\textbf{Average}} \\
\cmidrule(l{2pt}r{2pt}){2-5} \cmidrule(l{2pt}r{2pt}){6-9} \cmidrule(l{2pt}r{2pt}){10-13}
\textbf{Method} & DINO $\downarrow$ & FID $\downarrow$ & CLIP $\uparrow$ & mIoU $\uparrow$ & DINO $\downarrow$ & FID $\downarrow$ & CLIP $\uparrow$ & mIoU $\uparrow$ & DINO $\downarrow$ & FID $\downarrow$ & CLIP $\uparrow$ & mIoU $\uparrow$ \\
\midrule
CycleNet~\cite{xu2023cyclenet} & \underline{0.15} & 147.53 & 0.21 & \underline{0.35} & \underline{0.25} & 145.00 & 0.23 & \underline{0.51} & \underline{0.20} & 146.27 & 0.22 & \underline{0.43} \\
IP2P~\cite{brooks2022instructpix2pix} & \textbf{0.11} & 140.11 & 0.22 & \textbf{0.41} & \textbf{0.15} & 129.40 & \underline{0.26} & \textbf{0.60} & \textbf{0.13} & 134.76 & 0.24 & \textbf{0.51} \\
Qwen-Image~\cite{wu2025qwenimagetechnicalreport} & 0.50 & \underline{110.67} & \underline{0.26} & 0.33 & 1.14 & \underline{114.31} & \textbf{0.27} & 0.39 & 0.82 & \underline{112.49} & \underline{0.27} & 0.36 \\
DiffusionRenderer~\cite{DiffusionRenderer} & 1.57 & 191.42 & 0.25 & 0.17 & 1.34 & 143.50 & 0.24 & 0.11 & 1.46 & 167.46 & 0.25 & 0.14 \\
\midrule
\method~(ours) & 0.63 & \textbf{74.65} & \textbf{0.27} & 0.28 & 0.50 & \textbf{100.72} & \textbf{0.27} & 0.43 & 0.57 & \textbf{87.69} & \textbf{0.27} & 0.36 \\
\bottomrule
\end{tabular}
}
\caption{Quantitative results. \method~consistently outperforms all methods on FID and CLIP metrics for N$\to$D and D$\to$N translation, showing superiority in maintaining structural integrity while generating realistic lighting effects. CycleNet and IP2P often fail to perform significant relighting, their outputs are nearly identical to the source, which explains their best DINO and mIoU scores and very low FID and CLIP scores.}
\label{tab:quantitative_results}
\end{table}

\begin{figure*}[t]
    \centering
    \begin{subfigure}[b]{0.49\textwidth}
        \centering
        \includegraphics[width=\linewidth]{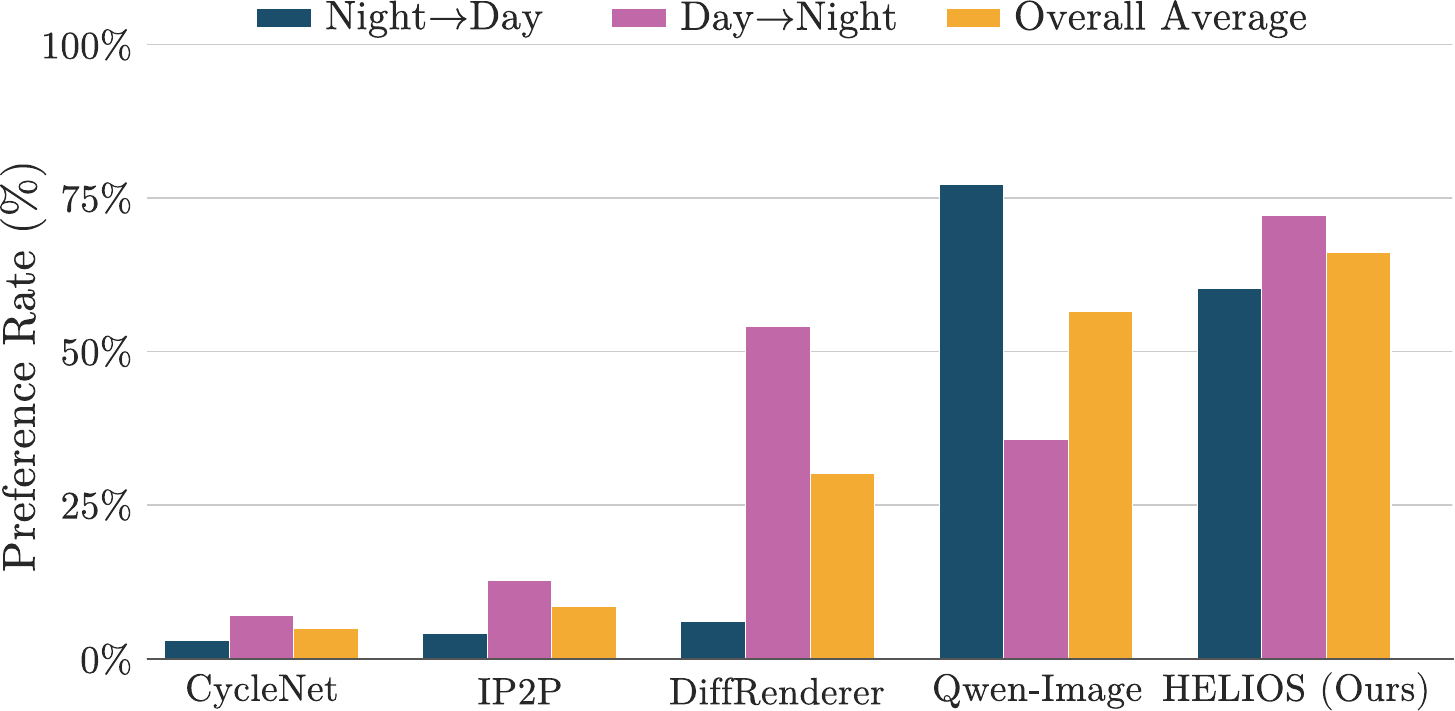}
        \caption{Preference Rate $\uparrow$}
        \label{fig:user_pref}
    \end{subfigure}%
    \hfill
    \begin{subfigure}[b]{0.49\textwidth}
        \centering
        \includegraphics[width=\linewidth]{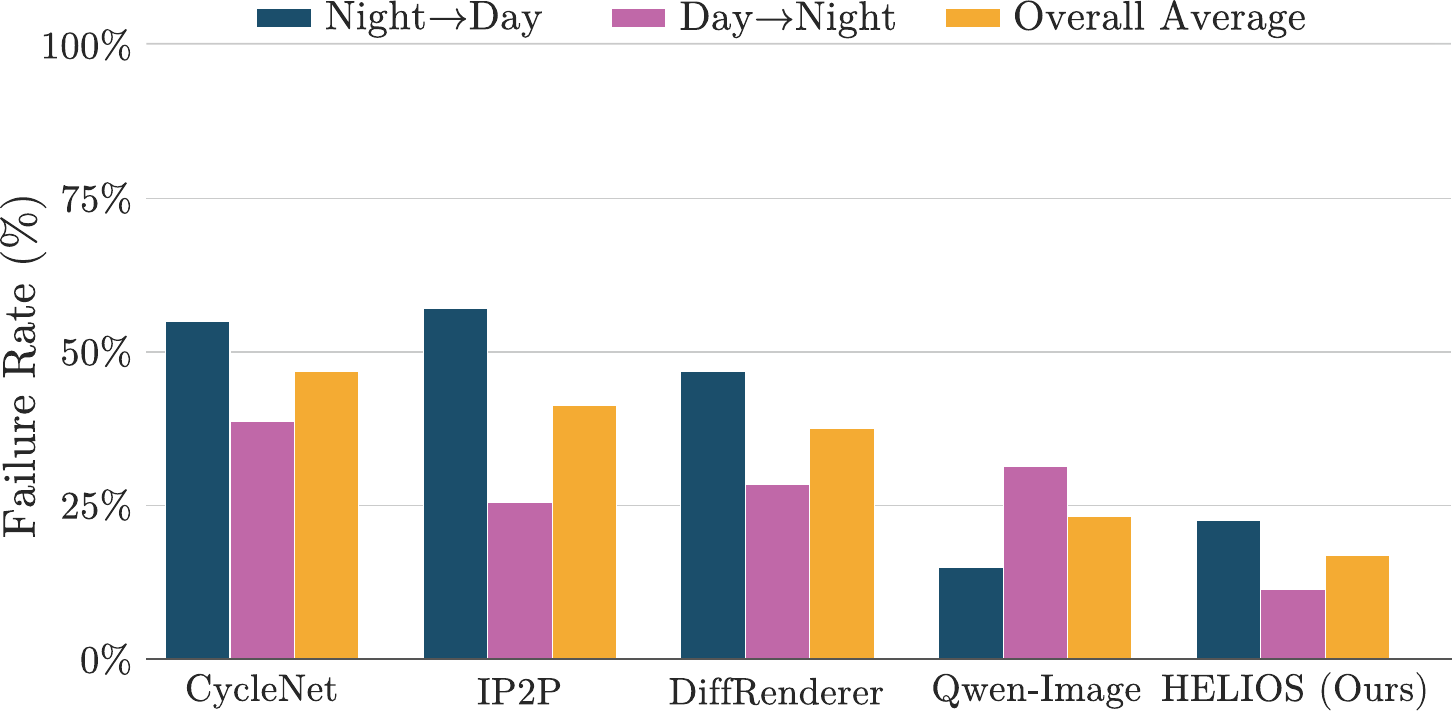}
        \caption{Failure Rate $\downarrow$}
        \label{fig:user_fail}
    \end{subfigure}
    
    \caption{\textbf{User Study Results.} We perform a user study to evaluate the performance of our method by (a) assessing the preference rate for daytime translation and (b) assessing the failure rate (non-realistic translation). \method~achieves the best results for Day$\to$Night translation and overall translation. Qwen-Image, being 20 times larger and 100 times more costly to run, achieves slightly better Day$\to$Night translation.}
    \label{fig:overall_user_study}
\end{figure*}

\subsection{Quantitative results}
We report in Tab~\ref{tab:quantitative_results} quantitative evaluation results for Night-to-Day, Day-to-Night and average on both tasks. 
Baseline methods such as CycleNet~\cite{xu2023cyclenet} and InstructPix2Pix~\cite{brooks2022instructpix2pix} achieve the best DINO and mIoU scores; however, this is primarily due to their failure to perform significant domain translation, copying the reference image. This lack of meaningful relighting is reflected in their low FID and CLIP scores.
DiffusionRenderer~\cite{DiffusionRenderer} reports the highest DINO and FID  scores and lower mIoU, reflecting a loss of structural integrity and a shift from the real data distribution. However, the model maintains a competitive CLIP score because the environment maps provide effective global lighting cues despite the geometric failures.
While Qwen-Image~\cite{wu2025qwenimagetechnicalreport} shows competitive results for Night-to-Day translation, it struggles with Day-to-Night tasks, failing to synthesize realistic nocturnal lighting while preserving scene structure. 
In contrast, our approach, \method, achieves the best balance across both tasks. It maintains a reasonably low DINO score, indicating high structural fidelity, while significantly outperforming baselines in FID and CLIP-Score, demonstrating superior distribution alignment and semantic relighting accuracy.

\subsection{User Study}
The user study evaluation involved 18 participants who performed 48 pairwise comparisons of images from a randomized set. For each trial, two randomly selected methods were compared against a ground-truth reference. If both methods failed to produce a realistic result, users had the option to select \textit{``both fail''}, which contributed to the failure rate score (we refer the reader to the supplementary materials for more details).

We summarize the human perceptual study results for both Night-to-Day and Day-to-Night tasks in Fig.~\ref{fig:overall_user_study}. 
The results reveal a significant performance asymmetry for Qwen-Image~\cite{wu2025qwenimagetechnicalreport}. While Qwen-Image shows impressive results in Night-to-Day translation, its performance collapses in the Day-to-Night task, where its win rate drops to 27.6\% and its failure rate doubles. Specifically, we observe that Qwen-Image frequently suffers from semantic hallucinations, such as incorrectly inserting non-existent vehicles or failing to maintain structural consistency. These failure cases are shown in the supplementary material. 
In Night-to-Day translation, InstructPix2Pix~\cite{brooks2022instructpix2pix} generally fails to produce a meaningful change. However, it performs better in Day-to-Night scenarios. This is because IP2P can adjust the global ambient lighting of a scene, but it fails to synthesize the local light sources. CycleNet~\cite{xu2023cyclenet}  consistently reports high failure rates because it cannot translate any visible lighting effect.
In contrast, \method~provides consistent and reliable results across both translation tasks. Averaged across both tasks, our method achieves a higher win rate (58.8\%) and the lowest overall failure rate (19.7\%), significantly outperforming Qwen-Image's average of 48.9\% and 26.0\% respectively. These results confirm that~\method~produces more stable relighting results.

\begin{figure*}[tb] 
    \centering
    \def\fgsize{0.187}
    \scriptsize
    \setlength{\tabcolsep}{0.0020\linewidth}
    \renewcommand{\arraystretch}{1.0}
    \begin{tabular}{l@{\hspace{2pt}}c@{\hspace{5pt}}cc@{\hspace{5pt}}cc}
         & Input & \multicolumn{2}{c}{w/o Albedo Distillation} & \multicolumn{2}{c}{Full model} \\ 
         
         \rotatebox{90}{Night$\to$Day} &
         \includegraphics[width=\fgsize\columnwidth]{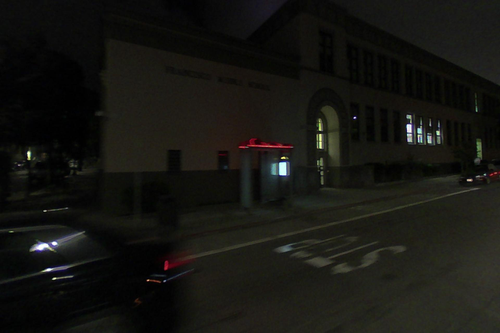} & 
         \includegraphics[width=\fgsize\columnwidth]{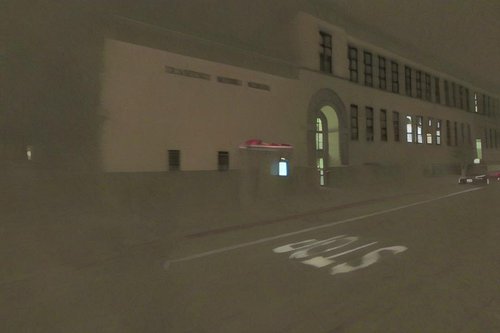} & 
         \includegraphics[width=\fgsize\columnwidth]{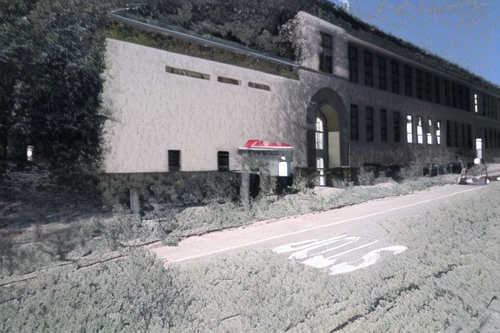} & 
         \includegraphics[width=\fgsize\columnwidth]{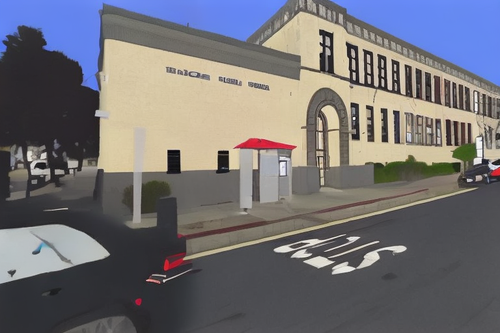} & 
         \includegraphics[width=\fgsize\columnwidth]{results_paper/Ours/040.png} \\[2pt]

         \rotatebox{90}{Day$\to$Night} &
         \includegraphics[width=\fgsize\columnwidth]{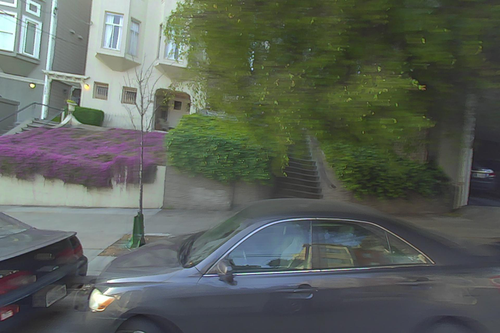} & 
         \includegraphics[width=\fgsize\columnwidth]{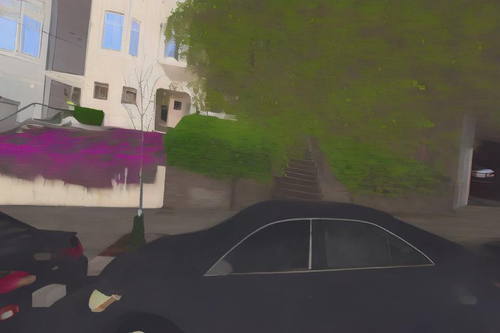} & 
         \includegraphics[width=\fgsize\columnwidth]{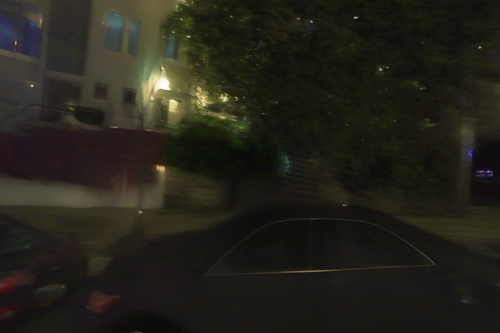} & 
         \includegraphics[width=\fgsize\columnwidth]{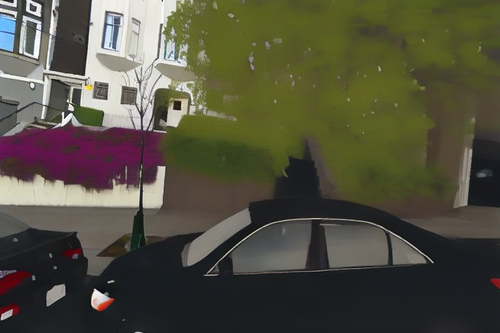} & 
         \includegraphics[width=\fgsize\columnwidth]{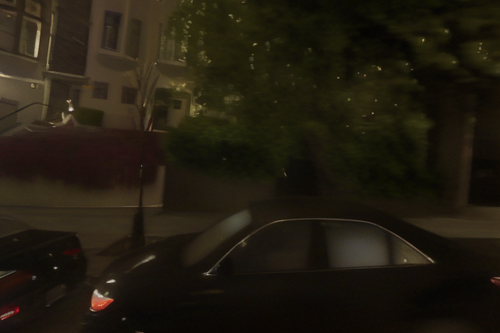} \\

         &  & Albedo & Output & Albedo & Output \\
        
    \end{tabular}
    \caption{\textbf{Ablation Study.} We compare our robust albedo estimator against a standard albedo estimator. For $N\to$D our albedo maintains accurate structures and colors in dark areas, providing a stable conditioning for high-fidelity relighting. The albedo computed with our robust estimator predicts sharper details for D$\to$N, which directly results in sharper relighting outputs.}
    \label{fig:albedo_ablation}
    
\end{figure*}

\begin{table}[tb]
\centering
\footnotesize
\setlength{\tabcolsep}{4pt}
\begin{tabular}{l cccc cccc}
\toprule
& \multicolumn{4}{c}{N$\to$D} & \multicolumn{4}{c}{D$\to$N} \\
\cmidrule(lr){2-5} \cmidrule(lr){6-9}
& DINO$\downarrow$ & FID$\downarrow$ & CLIP$\uparrow$ & mIoU$\uparrow$ & DINO$\downarrow$ & FID$\downarrow$ & CLIP$\uparrow$ & mIoU$\uparrow$ \\
\midrule
w/o distill. & 0.74 & 141.75 & 0.20 & 0.21 & 0.63 & 100.77 & 0.26 & 0.34 \\
Full HELIOS  & \textbf{0.63} & \textbf{74.65} & \textbf{0.27} & \textbf{0.28} & \textbf{0.50} & \textbf{100.72} & \textbf{0.27} & \textbf{0.43} \\
\bottomrule
\end{tabular}
\caption{Ablation: impact of albedo distillation.}
\label{tab:ablation_full}
\vspace{-7mm} 
\end{table}

\subsection{Ablation study}

We evaluate our robust albedo estimator in Fig. \ref{fig:albedo_ablation} and Tab. \ref{tab:ablation_full}, comparing it against a baseline using standard pre-computed albedo maps from DiffusionRender~\cite{DiffusionRenderer}. The results indicate that while basic albedo decomposition is sufficient for Day-to-Night (D$\to$N) translation, it performs poorly in Night-to-Day (N$\to$D) scenarios. Standard estimators often fail to recover geometry in regions obscured by deep shadows or low nighttime visibility, leading to structural artifacts in the relit output.
In contrast, our distilled cross-domain albedo estimator provides a more reliable structural prior. By leveraging knowledge from the daytime domain, the model recovers accurate scene geometry and synthesizes realistic daytime illumination even from degraded nighttime inputs. Furthermore, for D$\to$N translation, our distillation strategy produces significantly sharper albedo maps. This increased high-frequency detail directly translates to sharper relighting results.

\section{Discussion}
\subsection{Applications}

\begin{figure*}[t]
    \centering
    \def\fgsize{0.23}
    \scriptsize %
    \setlength{\tabcolsep}{1pt} %
    \renewcommand{\arraystretch}{0.8} %
    
    \begin{tabular}{cccc}

        \includegraphics[width=\fgsize\linewidth]{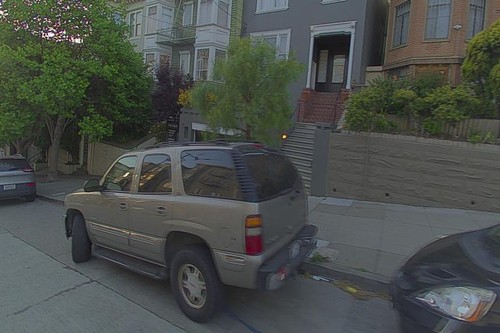} &
        \includegraphics[width=\fgsize\linewidth]{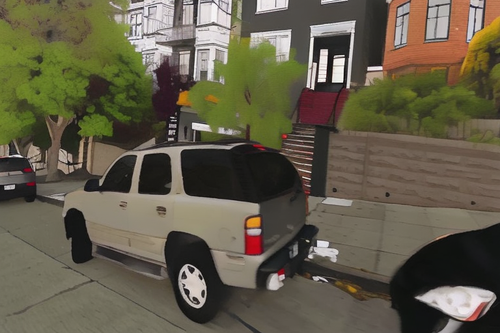} &
        \includegraphics[width=\fgsize\linewidth]{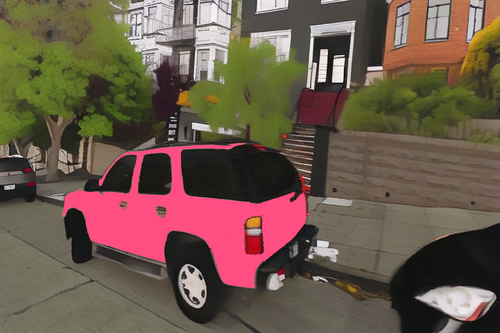} &
        \includegraphics[width=\fgsize\linewidth]{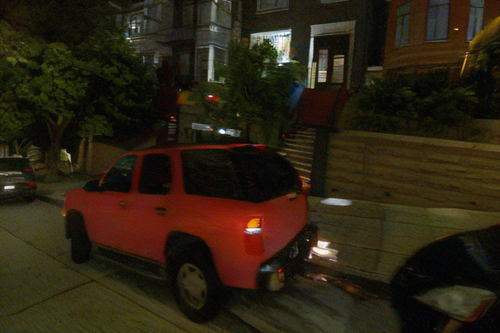} 
              \\ 
        
         Input   & Albedo & Edited albedo & Final edit \\
    \end{tabular}
    
    \vspace{-2mm} %
    \caption{\textbf{Application.} \method~enables color and lighting editing for data augmentation.}
    \label{fig:color-editing}
    \vspace{-2mm} 
\end{figure*}

\paragraph{Color editing.} \method~enables color editing for data augmentation. As shown in Fig.~\ref{fig:color-editing}, modifying the albedo color within a mask and relighting it using~\method~produces realistic scene edits. 

\begin{figure}
    \begin{minipage}[c]{.49\textwidth}
        \centering
        \setlength{\tabcolsep}{3pt}
        \begin{tabular}{cc}
            \includegraphics[width=0.47\linewidth]{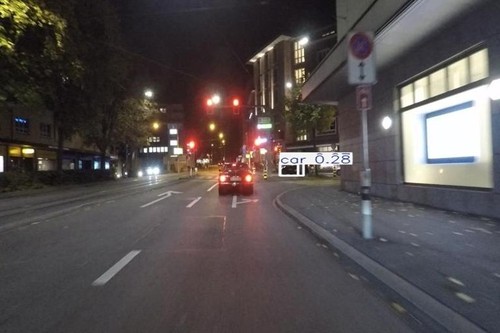} & \includegraphics[width=0.47\linewidth]{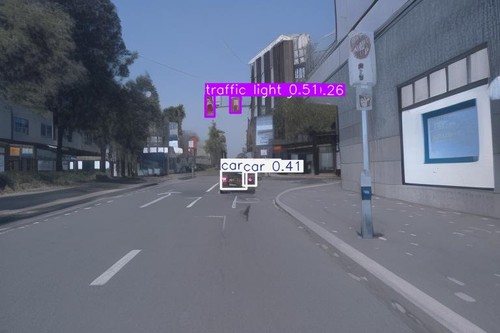} \\
            Input image & Night$\to$Day (ours)
        \end{tabular}
        \vspace{-0.3cm}
        \captionof{figure}{Downstream Object Detection Performance on ACDC dataset~\cite{acdc}.}
        \label{fig:yolo}
    \end{minipage}
    \hfill
    \begin{minipage}[c]{0.48\textwidth}

        \vspace{0.3cm}
        \centering
        \scriptsize
        \setlength{\tabcolsep}{3pt}
        \begin{tabular}{lcc}
            \toprule
            Method & mAP@50 & mAP@75 \\
            \midrule
            Night (input) & 6.0 & 2.7 \\
            CycleNet~\cite{xu2023cyclenet} & 5.0 & 2.7 \\
            IP2P~\cite{brooks2022instructpix2pix} & \underline{6.1} & \underline{2.8} \\
            Qwen-Image~\cite{wu2025qwenimagetechnicalreport} & 4.1 & 2.3 \\
            DiffusionRenderer~\cite{DiffusionRenderer} & fail & fail \\
            \midrule
            \method~(ours) & \textbf{10.0} & \textbf{8.0} \\
            \bottomrule 
            \multicolumn{3}{l}{\scriptsize Night baseline: mAP@50=6.0, mAP@75=2.7} \\
        \end{tabular}
                \caption{Comparison of YOLOv11 performance on night images and relit results.}
        \label{tab:yolo}
    \end{minipage}
    \vspace{-7mm} 
\end{figure}

\paragraph{Image enhancement.} We evaluate the robustness of our method for perception tasks on the ACDC dataset~\cite{acdc}. Using YOLOv11~\cite{Khanam2024YOLOv11AO} to report mAP on night scenes, we find that relighting to daytime improves detection performance by 5.3\% for  mAP@70 (Tab.~\ref{tab:yolo}). This confirms that \method~effectively recovers details that improve object detection in night conditions.

\subsection{Limitations} 
While \method~is robust to various outdoor lighting conditions, it remains limited by extremely low-light nocturnal environments where predicting accurate albedo-like images and relighting results becomes challenging (we refer the reader to the supplementary materials).

\subsection{Conclusion}
In this work, we presented \method, a method for unpaired image relighting in real-world driving scenarios. Trained without paired multi-illumination data or synthetic supervision, we introduced an albedo-based conditioning into a cycle-consistent diffusion pipeline to prevent identity collapse. We addressed the challenge of low-visibility nighttime conditions through an albedo distillation strategy that ensures structural preservation across both daytime and nighttime domains. Furthermore, we replaced coarse text prompts with a finer continuous control mechanism based on GPS-derived solar angles, enabling precise and smooth lighting manipulation. Quantitative and qualitative evaluations demonstrate that \method~outperforms state-of-the-art methods in preserving scene structure and generating realistic lighting.

\clearpage
\setcounter{page}{1}

\section{Dataset Distribution and Illumination Analysis}
\label{sec:distribution_analysis}

\begin{figure}[t]
  \centering
  \begin{subfigure}[b]{0.48\linewidth}
    \centering
    \includegraphics[width=\textwidth]{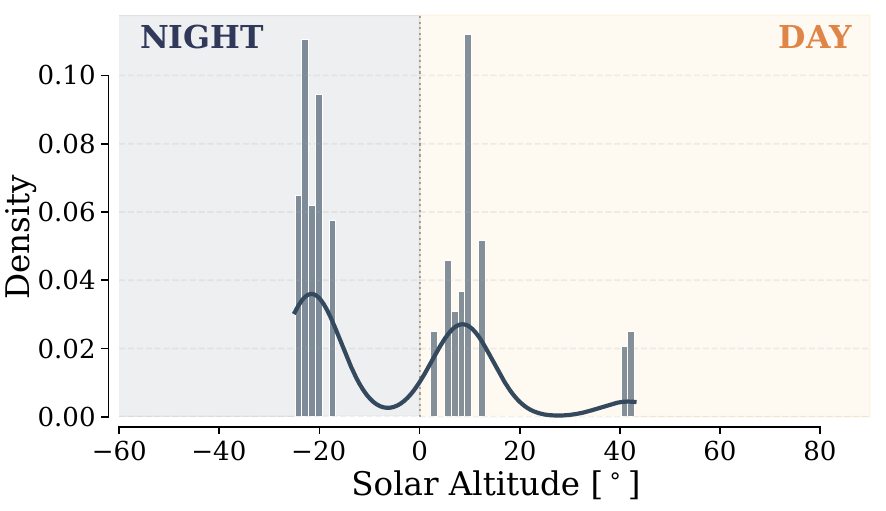}
    \caption{Pandaset}
    \label{fig:dist_pandaset}
  \end{subfigure}
  \hfill
  \begin{subfigure}[b]{0.48\linewidth}
    \centering
    \includegraphics[width=\textwidth]{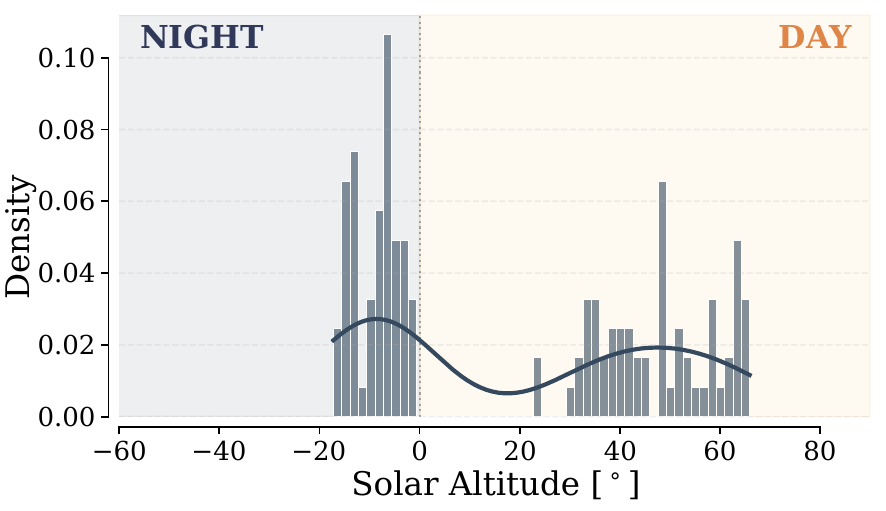}
    \caption{nuScenes}
    \label{fig:dist_nuscenes}
  \end{subfigure}

  \vspace{1em} %

  \begin{subfigure}[b]{0.48\linewidth} %
    \centering
    \includegraphics[width=\textwidth]{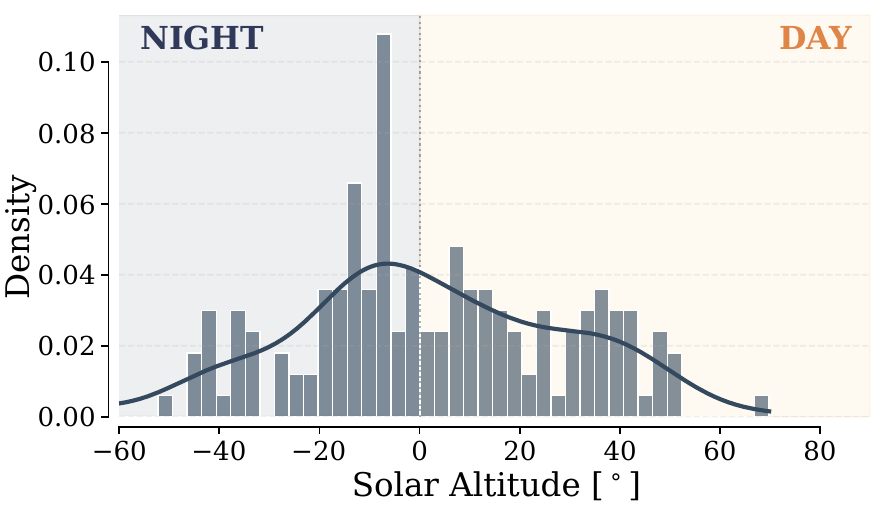}
    \caption{Waymo Open Dataset}
    \label{fig:dist_waymo}
  \end{subfigure}

    \caption{\textbf{Solar altitude distribution across training datasets.} We plot the density of samples relative to the sun's elevation.}
  \label{fig:dataset_comparison}
\end{figure}

Fig.~\ref{fig:dataset_comparison} illustrates the solar altitude distributions for Pandaset~\cite{pandaset}, nuScenes~\cite{nuscenes}, and Waymo~\cite{Waymo}. We observe that:

\begin{itemize}
    \item \textbf{Pandaset:} Solar altitudes are highly clustered with sharp peaks at specific angles. This indicates that the data was recorded during a small number of driving sessions at fixed times of day.
    \item \textbf{nuScenes:} This dataset shows a wider range of daytime coverage, specifically between $20^\circ$ and $65^\circ$. This variation is due to data collection in diverse locations, such as Singapore and Boston, which have different solar paths.
    \item \textbf{Waymo Open Dataset:} Waymo has the most continuous distribution, covering angles from deep night ($-40^\circ$) to high noon ($+70^\circ$). The smoother density curve shows that data was captured over a much larger number of sessions and a broader timeframe.
\end{itemize}

\noindent By training on these diverse sequences and sun elevations, \method~provides continuous relighting control for outdoor driving images. We refer to the video provided in the supplementary material for examples of controllable, smooth, and continuous relighting.

\section{User study details}

\label{sec:user_study}

 We compared our model against four state-of-the-art baselines across the two translation tasks: Day-to-Night (D2N) and Night-to-Day (N2D).

\subsection{Experimental Setup}

The study was implemented as a web-based interface. Each of the 18 participants performed a total of 48 randomized pairwise comparisons (24 rounds for each task). In each trial, a reference image was displayed alongside two randomly selected results from different methods.

To eliminate experimental bias, we ensured a randomized protocol: 

\begin{itemize}
    \item Image Selection: The reference image was randomly selected from a set of 1000 images.
    \item Method Pairing: For every comparison, we randomly paired two of the five methods, ensuring that over the course of the study, all methods were compared against one another in a statistically balanced manner.
    \item The side-by-side positioning (Left vs. Right) of the methods was randomized to prevent "side-bias" during the voting process.
\end{itemize}

\noindent Each trial presented the following instruction:
\begin{quote}
    \textit{Which of the following options represents the most realistic [Night/Day] version of the reference image while preserving its original structure?}
\end{quote}

A ``Both Failed'' option was provided to identify cases where neither result was satisfactory, usually due to identity loss or failed relighting.

\subsection{Evaluation Metrics}
We define two primary metrics to assess the model's performance:

\begin{itemize}
    \item \textbf{Win Rate (\%):} This represents the \textbf{preference rate}. It is the percentage of trials in which a specific method's output was chosen as superior to its competitor. A higher win rate indicates greater visual realism and better preservation of the scene's identity.
    \item \textbf{Fail Rate (\%):} It reflects the proportion of failure case of the model, where the translation task cannot be done successfully either by producing an image with too many unrealistic artifacts or by not following the relighting instruction.
\end{itemize}

\section{Additional results}

\begin{figure*}[tb] 
\centering
\def\fgsize{0.16}
\scriptsize
\setlength{\tabcolsep}{0.0015\linewidth}
\renewcommand{\arraystretch}{1.0}
\begin{tabular}{ccccccc}%
          Input & CycleNet~\cite{xu2023cyclenet} &IP2P~\cite{brooks2022instructpix2pix} & Qwen-Img~\cite{wu2025qwenimagetechnicalreport} & DiffRendr~\cite{DiffusionRenderer} &\method~(ours)  \\ 

    \includegraphics[clip=false, trim={0 0 0 0},width=\fgsize\columnwidth]{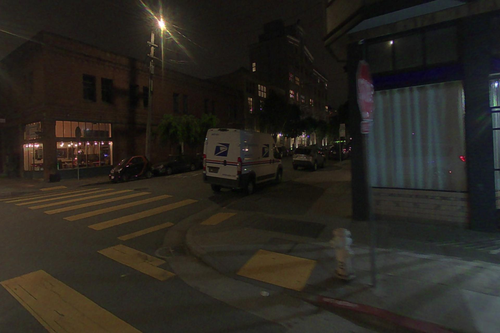} & 
    \includegraphics[clip=false, trim={0 0 0 0},width=\fgsize\columnwidth]{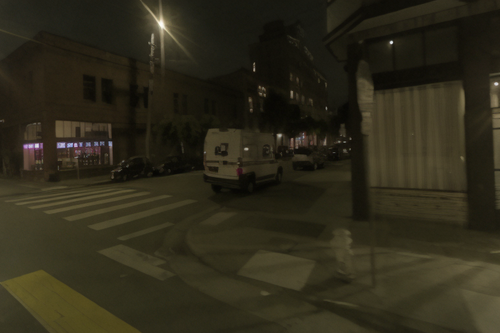} & 
    \includegraphics[clip=false, trim={0 0 0 0},width=\fgsize\columnwidth]{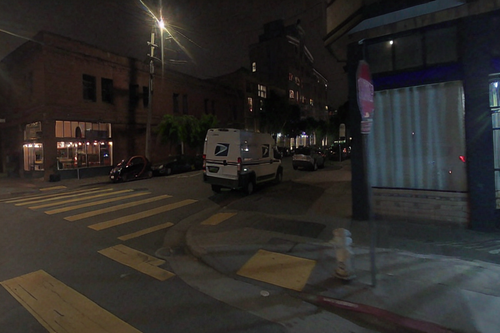} & 
    \includegraphics[clip=false, trim={0 0 0 0},width=\fgsize\columnwidth]{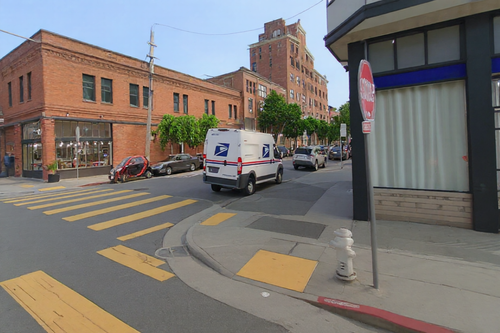} & 
    \includegraphics[clip=false, trim={0 0 0 0},width=\fgsize\columnwidth]{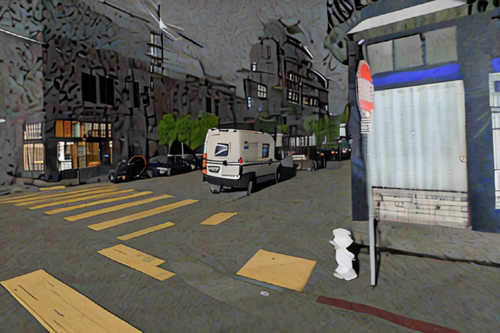} & 
    \includegraphics[clip=false, trim={0 0 0 0},width=\fgsize\columnwidth]{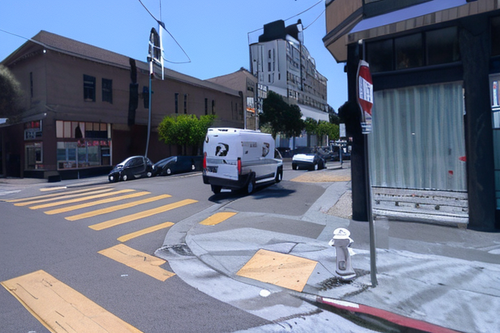} &  \\
    
    \includegraphics[clip=false, trim={0 0 0 0},width=\fgsize\columnwidth]{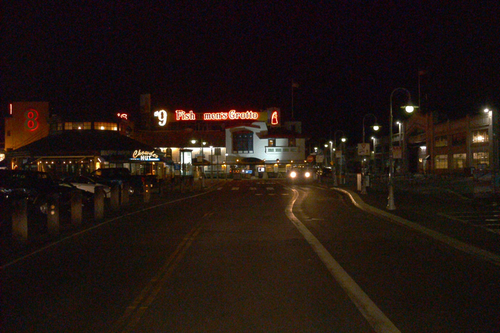} & 
    \includegraphics[clip=false, trim={0 0 0 0},width=\fgsize\columnwidth]{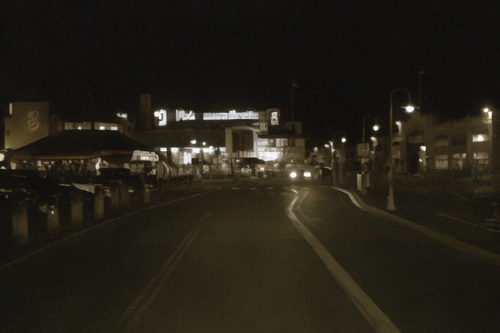} & 
    \includegraphics[clip=false, trim={0 0 0 0},,width=\fgsize\columnwidth]{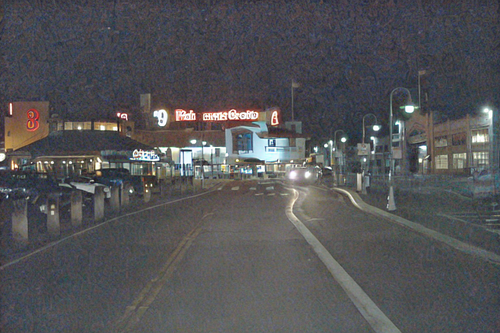} & 
    \includegraphics[clip=false, trim={0 0 0 0},,width=\fgsize\columnwidth]{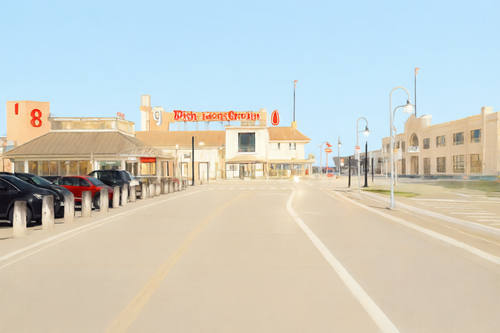} & 
     \includegraphics[clip=false, trim={0 0 0 0},,width=\fgsize\columnwidth]{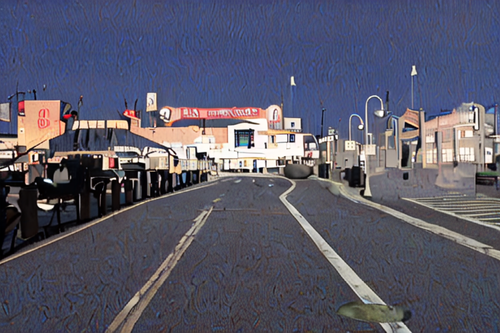} & 
    \includegraphics[clip=false, trim={0 0 0 0},,width=\fgsize\columnwidth]{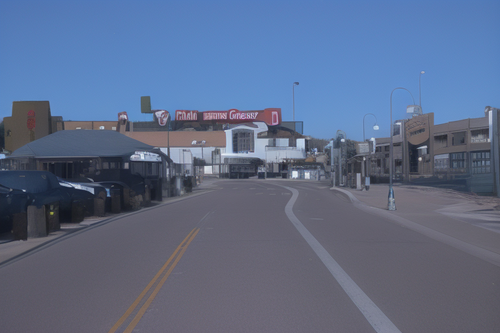} & \\
    
    \includegraphics[clip=false, trim={0 0 0 0},width=\fgsize\columnwidth]{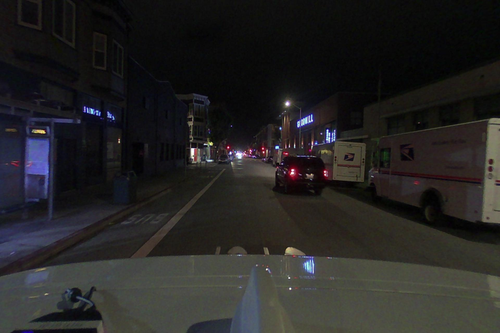} & 
    \includegraphics[clip=false, trim={0 0 0 0},width=\fgsize\columnwidth]{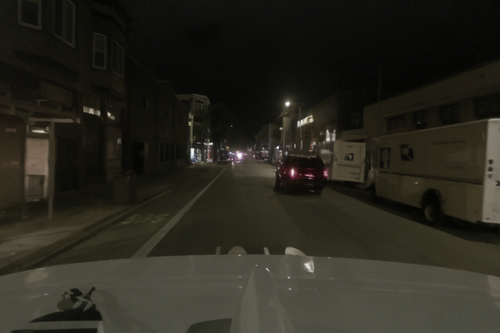} & 
    \includegraphics[clip=false, trim={0 0 0 0},width=\fgsize\columnwidth]{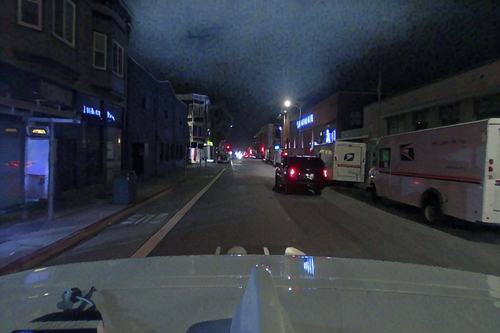} & 
    \includegraphics[clip=false, trim={0 0 0 0},width=\fgsize\columnwidth]{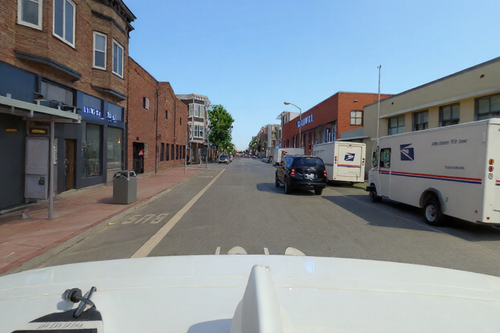} & 
    \includegraphics[clip=false, trim={0 0 0 0},width=\fgsize\columnwidth]{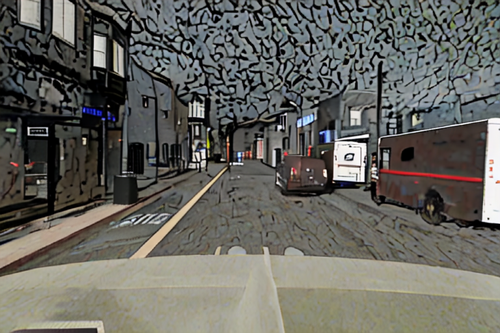} & 
    \includegraphics[clip=false, trim={0 0 0 0},width=\fgsize\columnwidth]{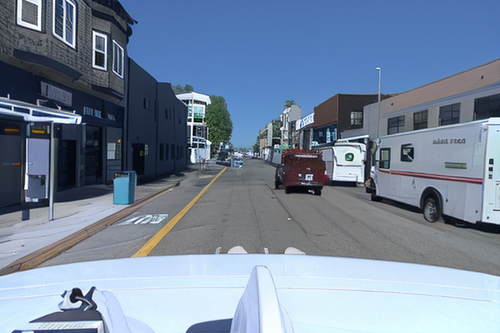} &  \\
        
    \includegraphics[clip=false, trim={0 0 0 0},width=\fgsize\columnwidth]{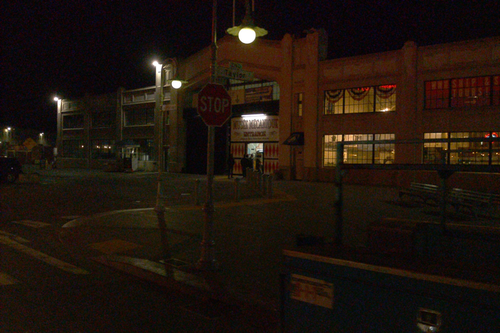} & 
    \includegraphics[clip=false, trim={0 0 0 0},width=\fgsize\columnwidth]{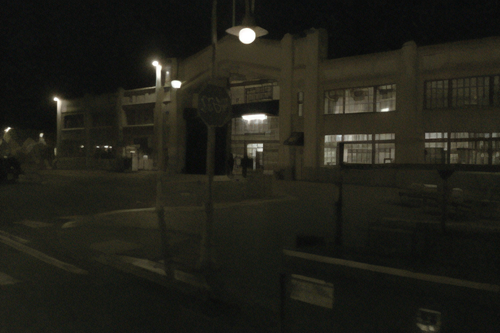} & 
    \includegraphics[clip=false, trim={0 0 0 0},width=\fgsize\columnwidth]{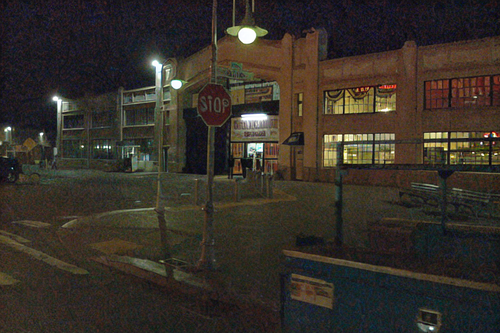} & 
    \includegraphics[clip=false, trim={0 0 0 0},width=\fgsize\columnwidth]{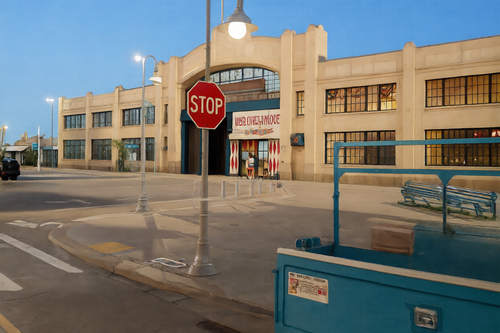} & 
    \includegraphics[clip=false, trim={0 0 0 0},width=\fgsize\columnwidth]{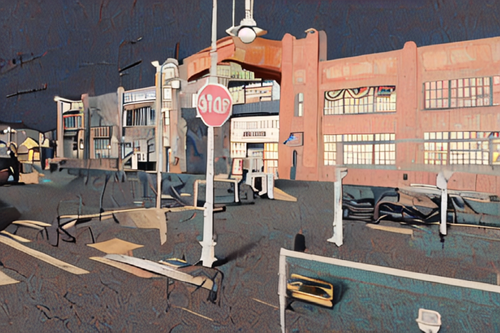} & 
    \includegraphics[clip=false, trim={0 0 0 0},width=\fgsize\columnwidth]{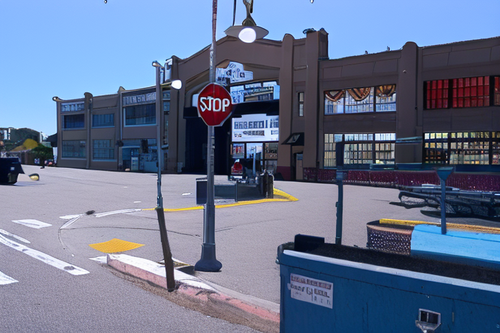} &  \\
\end{tabular}
\caption{We qualitatively compare the re-lit images. We show that~\method~predicts realistic Night-to-Day relighting.}
\label{fig:add-N2D}

\end{figure*}

\begin{figure*}[tb] 
\centering
\def\fgsize{0.16}
\scriptsize
\setlength{\tabcolsep}{0.0015\linewidth}
\renewcommand{\arraystretch}{1.0}
\begin{tabular}{ccccccc}%
          Input & CycleNet~\cite{xu2023cyclenet} &IP2P~\cite{brooks2022instructpix2pix} & Qwen-Img~\cite{wu2025qwenimagetechnicalreport} & DiffRendr~\cite{DiffusionRenderer} &\method~(ours)  \\ 

    \includegraphics[clip=false, trim={0 0 0 0},width=\fgsize\columnwidth]{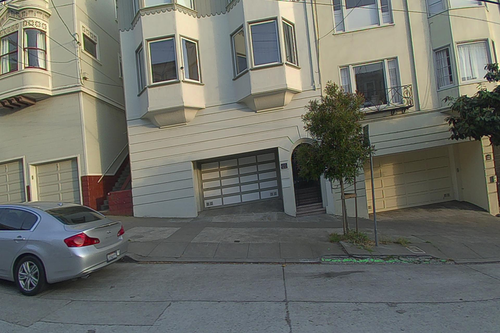} & 
    \includegraphics[clip=false, trim={0 0 0 0},width=\fgsize\columnwidth]{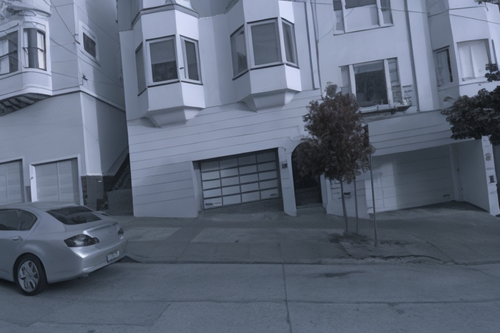} & 
    \includegraphics[clip=false, trim={0 0 0 0},width=\fgsize\columnwidth]{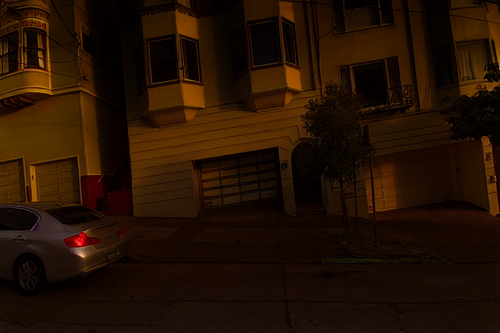} & 
    \includegraphics[clip=false, trim={0 0 0 0},width=\fgsize\columnwidth]{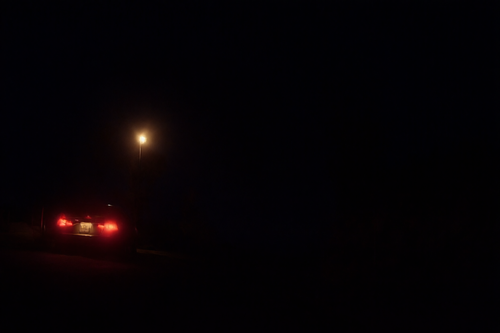} & 
    \includegraphics[clip=false, trim={0 0 0 0},width=\fgsize\columnwidth]{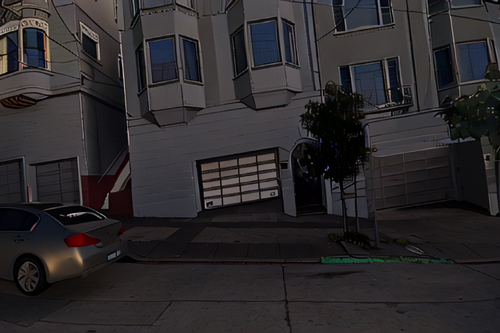} & 
    \includegraphics[clip=false, trim={0 0 0 0},width=\fgsize\columnwidth]{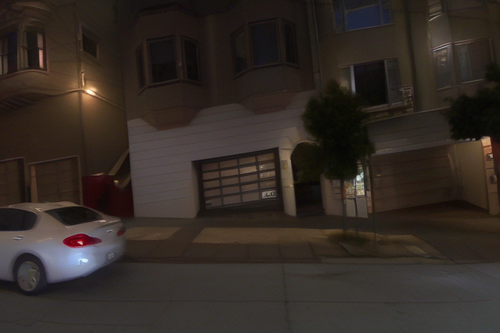} &  \\
    
    \includegraphics[clip=false, trim={0 0 0 0},width=\fgsize\columnwidth]{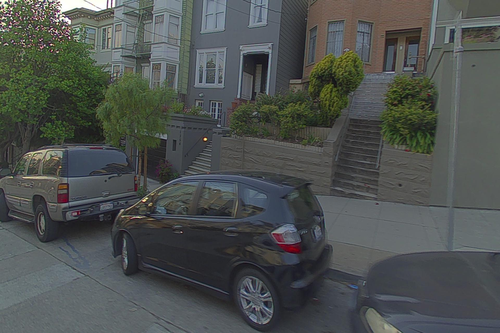} & 
    \includegraphics[clip=false, trim={0 0 0 0},width=\fgsize\columnwidth]{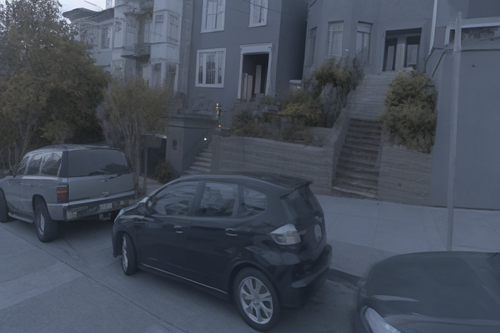} & 
    \includegraphics[clip=false, trim={0 0 0 0},width=\fgsize\columnwidth]{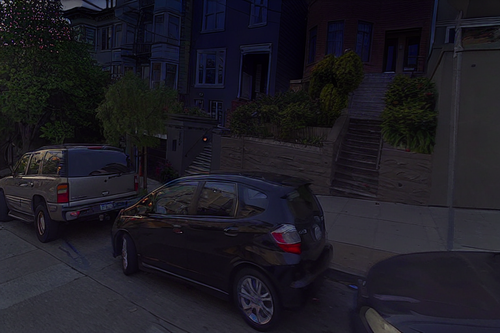} & 
    \includegraphics[clip=false, trim={0 0 0 0},width=\fgsize\columnwidth]{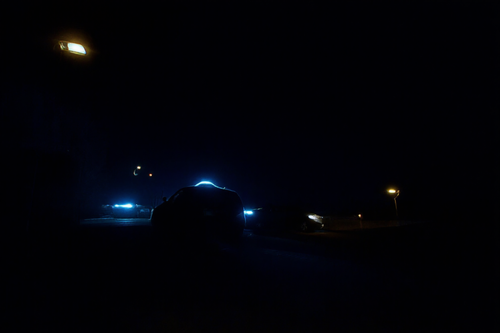} & 
    \includegraphics[clip=false, trim={0 0 0 0},width=\fgsize\columnwidth]{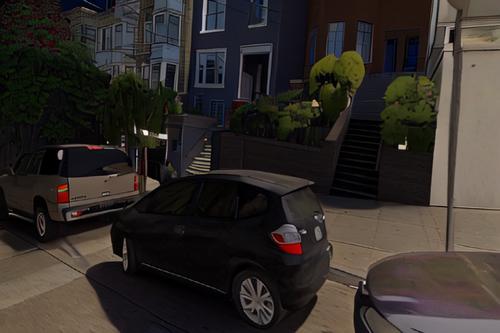} & 
    \includegraphics[clip=false, trim={0 0 0 0},width=\fgsize\columnwidth]{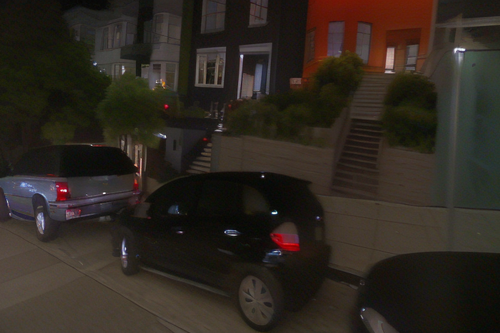} &  \\
    
    \includegraphics[clip=false, trim={0 0 0 0},width=\fgsize\columnwidth]{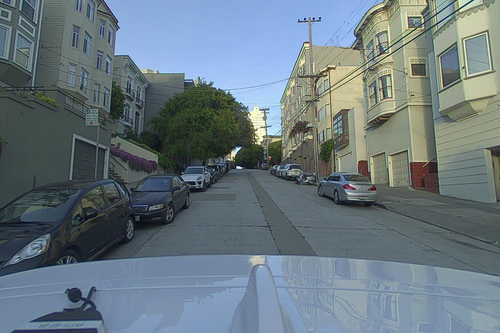} & 
    \includegraphics[clip=false, trim={0 0 0 0},width=\fgsize\columnwidth]{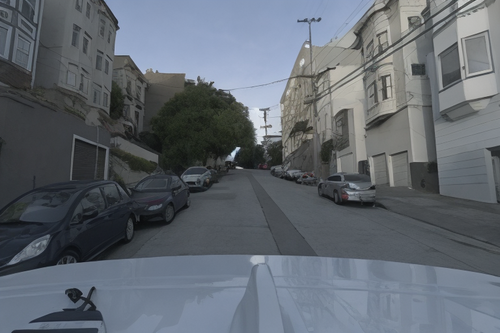} & 
    \includegraphics[clip=false, trim={0 0 0 0},width=\fgsize\columnwidth]{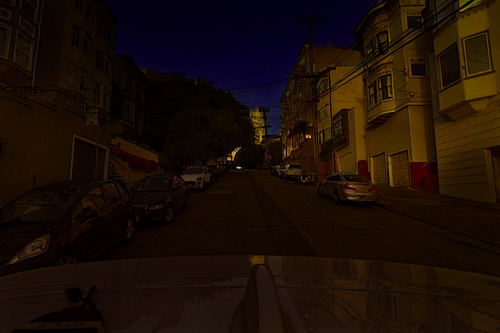} & 
    \includegraphics[clip=false, trim={0 0 0 0},width=\fgsize\columnwidth]{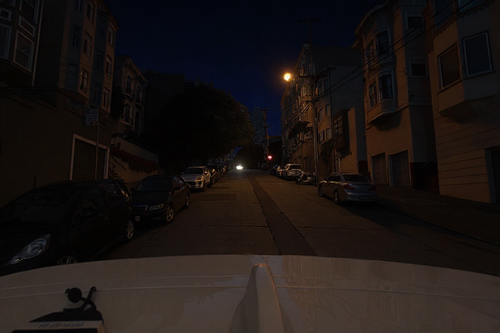} & 
    \includegraphics[clip=false, trim={0 0 0 0},width=\fgsize\columnwidth]{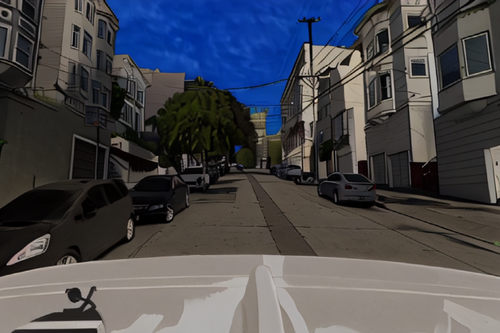} & 
    \includegraphics[clip=false, trim={0 0 0 0},width=\fgsize\columnwidth]{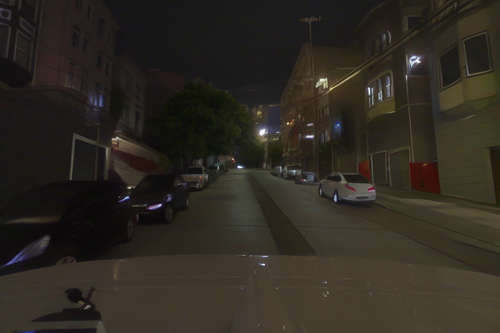} &  \\
        
\end{tabular}
\caption{We qualitatively compare the re-lit images. We show that~\method~predicts realistic Day-to-Night relighting.}
\label{fig:add-D2N}

\end{figure*}

We provide additional relighting results in Fig.~\ref{fig:add-N2D} and~\ref{fig:add-D2N}, for the two translation tasks. The results demonstrate the robustness of~\method. As shown, the model consistently preserves scene geometry while accurately modulating global illumination according to the target solar altitude.

\subsection{Qwen-Image failures} 
\begin{figure*}[tb] 
\centering
\def\fgsize{0.40}
\scriptsize
\setlength{\tabcolsep}{0.0015\linewidth}
\renewcommand{\arraystretch}{1.0}
\begin{tabular}{cc}%
          Input & Qwen-Img~\cite{wu2025qwenimagetechnicalreport}  \\ 

    \includegraphics[clip=false, trim={0 0 0 0},width=\fgsize\columnwidth]{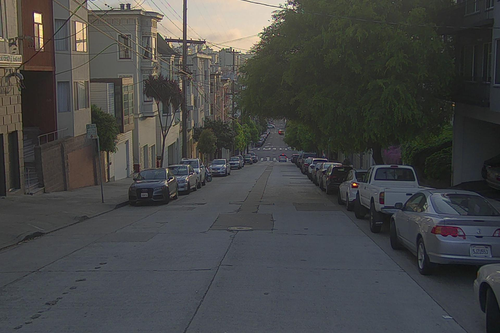} & 

    \includegraphics[clip=false, trim={0 0 0 0},width=\fgsize\columnwidth]{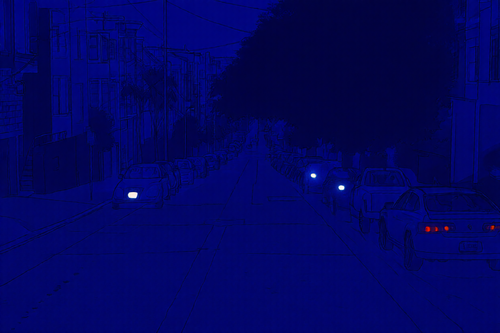} \\

        \includegraphics[clip=false, trim={0 0 0 0},width=\fgsize\columnwidth]{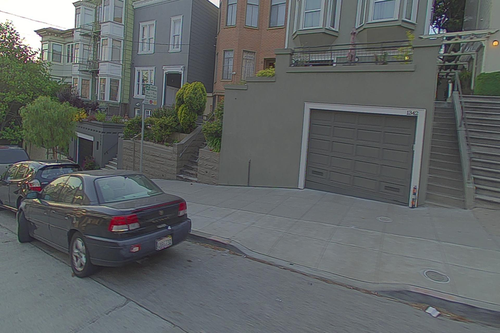} & 

    \includegraphics[clip=false, trim={0 0 0 0},width=\fgsize\columnwidth]{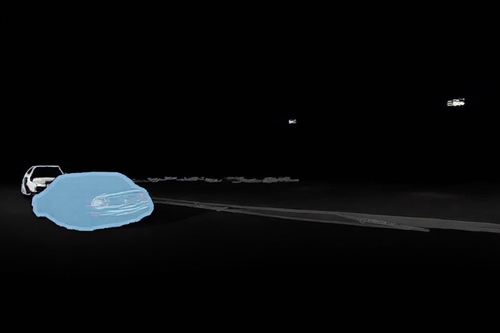}  \\

\end{tabular}
\caption{Failure of Qwen-Image~\cite{wu2025qwenimagetechnicalreport}} 
\label{fig:qwen-failure}

\end{figure*}

We show in Fig.~\ref{fig:qwen-failure} an example of a failure case of Qwen-Image~\cite{wu2025qwenimagetechnicalreport}. The baseline often exhibits modal instability due to its multi-task training. It occasionally confuses the relighting instruction with other tasks.

\subsection{Limitation}

\begin{figure*}[b] 
\centering
\def\fgsize{0.40}
\scriptsize
\setlength{\tabcolsep}{0.0015\linewidth}
\renewcommand{\arraystretch}{1.0}
\begin{tabular}{cc}%
          Input & \method~(ours) \\ 

    \includegraphics[clip=false, trim={0 0 0 0},width=\fgsize\columnwidth]{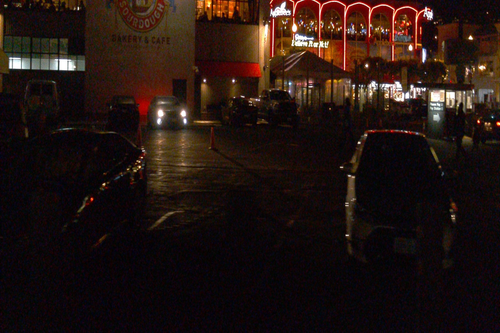} & 

    \includegraphics[clip=false, trim={0 0 0 0},width=\fgsize\columnwidth]{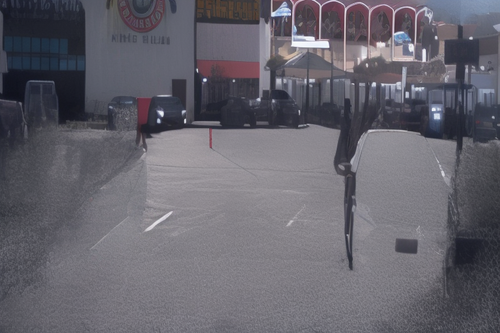} \\

\end{tabular}
\caption{Failure case of~\method~for Night-to-Day relighting task.} 
\label{fig:helios-failure}

\end{figure*}

Fig.~\ref{fig:helios-failure} shows a failure case for \method. In images with extreme nighttime lighting, the model fails to recover structural details. This limitation could be addressed by fine-tuning a more robust albedo estimator trained on a wider distribution of extremely low-light data to better capture structural details.

\newpage

\clearpage

\bibliographystyle{splncs04}
\bibliography{main}
\end{document}